\documentclass[10pt,journal,compsoc]{IEEEtran}

\usepackage{booktabs,multirow,siunitx,xcolor,colortbl}
\usepackage{graphicx}
\usepackage[cmex10]{amsmath}
\usepackage{amssymb}
\usepackage{algorithm}
\usepackage{algorithmic}
\usepackage{cite}
\usepackage{bm}
\usepackage{mathtools}
\usepackage{subfigure}
\usepackage{url}
\usepackage{hyperref}
\usepackage{ragged2e}
\newcommand{\argmin}{\operatornamewithlimits{argmin}}
\newcommand{\argmax}{\operatornamewithlimits{argmax}}

\newcommand{\bx}{{\mathbf{x}}}
\newcommand{\by}{{\mathbf{y}}}
\newcommand{\bz}{{\mathbf{z}}}
\newcommand{\bu}{{\mathbf{u}}}
\newcommand{\bw}{{\mathbf{w}}}

\allowdisplaybreaks[4]
\definecolor{rowgray}{gray}{0.93}

\newtheorem{remark}{Remark}

\begin{document}

\title{Scalable Joint Resource Allocation for SLO-Constrained LLM Inference in Heterogeneous GPU Clouds}

\author{\IEEEauthorblockN{Jiaming~Cheng,~\IEEEmembership{Member,~IEEE}, and Duong~Tung~Nguyen,~\IEEEmembership{Member,~IEEE}}
\thanks{The authors are with the Ira A. Fulton Schools of Engineering, Arizona State University, Tempe, AZ, USA. Email: \{jiaming, duongnt\}@asu.edu.}}

\IEEEtitleabstractindextext{%
\begin{abstract}
\justifying
Serving large language model (LLM) inference in cloud environments requires jointly optimizing model selection, heterogeneous GPU provisioning, parallelism configuration, and workload routing under tightly coupled latency, accuracy, memory, and budget constraints. While mixed-integer linear programming (MILP) formulations can capture this joint allocation problem, their computational cost limits their use for frequent re-optimization under demand variability. Existing heuristics, in contrast, typically optimize placement, provisioning, parallelism, or routing independently and often produce infeasible solutions when system-level constraint coupling is enforced. This paper presents a scalable framework for SLO-constrained LLM inference based on joint resource allocation. We formulate the problem as an MILP with a two-phase delay model that captures both prefill and autoregressive decoding under tensor and pipeline parallelism.
To address this problem, we develop two constraint-aware heuristics: a \textit{Greedy Heuristic} (GH) for single-pass allocation and an \textit{Adaptive Greedy Heuristic} (AGH) that enhances GH with multi-start construction, local search, and GPU consolidation. These methods explicitly enforce feasibility through parallelism-aware filtering, effective cost-based ranking, and adaptive parallelism scaling on active resources. Experiments using workloads calibrated to the Azure LLM Inference Trace show that GH produces feasible solutions within one second, while AGH achieves near-optimal performance within three seconds, scaling to large instances where exact solvers fail to converge. Under out-of-sample stress with up to 1.5$\times$ delay and accuracy inflation, AGH degrades gracefully through its provisioned headroom, keeping cost and SLO violations far lower, while cost-minimal MILP solutions degrade significantly. Across a synthetic volatility study and a real Azure diurnal trace, AGH’s headroom-aware allocations remain far cheaper than the exact MILP while maintaining SLO compliance. The results further show that robustness is driven primarily by the allocation itself rather than frequent re-optimization, as the static AGH plan absorbs substantial demand variability while the sub-three-second runtime provides an additional mechanism for adapting to non-stationary demand shifts.
\end{abstract}

}

\maketitle
\IEEEdisplaynontitleabstractindextext
\IEEEpeerreviewmaketitle

\section{Introduction}

Large language models (LLMs) have become a core component of modern cloud services, enabling applications such as conversational agents, code generation, translation, and multimodal content creation. These services rely on large-scale inference pipelines that must deliver responses within strict service-level objectives (SLOs) on latency and output quality. As model sizes continue to grow, often reaching tens or hundreds of billions of parameters, the cost of serving LLM inference on heterogeneous GPU infrastructure has become a major operational challenge for cloud providers~\cite{chien2023reducing}.

A key characteristic of modern LLM serving is that large models typically do not fit on a single GPU and must be distributed across multiple devices. Two complementary parallelism strategies are commonly used. \textit{Tensor parallelism} (TP) partitions model weights across co-located GPUs, reducing per-device memory and compute requirements while introducing inter-GPU communication overhead during decoding. \textit{Pipeline parallelism} (PP) distributes model layers across sequential stages, enabling larger models to be served at the cost of pipeline inefficiencies. 
In addition, production clusters employ heterogeneous GPU tiers (e.g., A6000, A100, H100) and multiple numerical precisions (e.g., FP16, INT8, INT4)
~\cite{stojkovic2024dynamollm,patel2024splitwise} to balance cost, performance, and accuracy. The combination of model choices, hardware tiers, precision levels, and parallelism configurations creates a large deployment space with thousands of feasible options, only a subset of which satisfy latency, accuracy, and cost constraints.

This heterogeneity creates a tightly coupled resource allocation problem spanning deployment, provisioning, parallelism configuration, and workload routing. A service provider must jointly determine which models to deploy, which GPU tiers to provision, how to configure tensor and pipeline parallelism, and how to distribute workload across deployed resources. These decisions are governed by shared system constraints, including per-GPU memory after model partitioning (sharding), compute throughput, end-to-end latency, quantization-induced accuracy degradation, storage capacity, and a global cost budget. Consequently, decisions that appear optimal in isolation may become infeasible when evaluated under system-level constraints. This paper develops a scalable framework for jointly optimizing these decisions under dynamic demand variability while preserving latency, accuracy, and budget guarantees.

These tradeoffs affect both cloud providers and end users.
Overprovisioning wastes budget on underutilized resources, whereas underprovisioning leads to SLO violations and unmet demand. End users experience the resulting degradation through increased response latency and reduced output quality caused by aggressive quantization or insufficient compute capacity. 
At the same time, hardware and precision choices influence efficiency: smaller quantized models deployed on appropriate GPU tiers can serve the same workload with substantially lower resource consumption than a full-precision configuration. 

Existing work addresses different aspects of this problem but does not provide a scalable solution for joint resource allocation under tightly coupled system constraints. 
High-throughput LLM serving systems improve execution efficiency through techniques such as paged-attention key--value cache management~\cite{kwon2023vllm}, prefill--decode disaggregation~\cite{zhong2024distserve}, and locality-aware cross-region load balancing~\cite{xia2025skylb}, but assume fixed deployment configurations. Optimization-based approaches formulate heterogeneous placement and routing as mixed-integer optimization problems~\cite{mei2025helix,jiang2025demystify,cheng_greenllm}, but their computational complexity grows rapidly with problem size, limiting scalability. Heuristic approaches~\cite{stojkovic2024dynamollm,kim2025cost,zhao2025seallm} achieve faster runtime but typically optimize placement, provisioning, parallelism, and routing independently, often producing infeasible allocations when memory, latency, accuracy, and budget constraints interact.

To address this challenge, we present a scalable framework for joint resource allocation in heterogeneous GPU clouds for SLO-constrained LLM inference. We formulate the problem as a mixed-integer linear program (MILP) with a two-phase delay model that captures both prefill and autoregressive decoding under tensor and pipeline parallelism. To achieve scalable real-time operation, we develop two constraint-aware heuristics: a Greedy Heuristic (GH) for single-pass allocation and an Adaptive Greedy Heuristic (AGH) that augments GH with multi-start construction, local search, and GPU consolidation. 

The proposed framework adopts a feasibility-first allocation strategy in which candidate configurations are filtered and ranked according to their ability to satisfy memory, latency, and accuracy constraints before cost is considered. The proposed heuristics maintain valid allocations through three mechanisms: parallelism-aware filtering, effective cost-based ranking, and adaptive parallelism scaling on active resources. Together, these mechanisms bridge the gap between exact optimization and fast heuristic planning, enabling scalable joint decision-making for production-scale LLM serving systems.

\textit{A key novelty of this work is the integration of joint parallelism configuration, heterogeneous GPU provisioning, and constraint-aware heuristic allocation in a single, seconds-scale allocation loop}. Unlike conventional heuristics that optimize individual decision dimensions independently, the proposed framework explicitly accounts for interacting memory, latency, accuracy, storage, and budget constraints while maintaining runtime suitable for operational re-optimization.
The experimental results further reveal that the primary source of robustness is not frequent re-optimization itself, but the headroom-aware allocation produced by AGH, which absorbs substantial demand variability even when deployments remain fixed. Our contributions are summarized as follows:

\begin{itemize}
\setlength{\itemsep}{2pt}
\item \textbf{Unified optimization framework:} We formulate joint model selection, heterogeneous GPU provisioning, tensor/pipeline parallelism configuration, and workload routing as an MILP for SLO-constrained LLM inference with a two-phase delay model.
\item \textbf{Constraint-aware heuristic design:} We develop scalable heuristics that preserve feasibility under coupled memory, latency, accuracy, storage, and budget constraints through parallelism-aware filtering and adaptive resource scaling.

\item \textbf{Numerical results:} On workloads calibrated to the Azure LLM Inference Trace, we benchmark GH and AGH against the exact MILP optimum and three state-of-the-art-derived heuristic baselines (an OR LP-relaxation, a decoupled VM-then-routing scheme after Kim~et~al., and a homogeneous-fleet scheme after DynamoLLM). The proposed methods match the exact MILP optimum within a few percent on instances the solver completes, and return feasible plans within seconds with over $260\times$ speedup on large-scale instances where the exact solver fails to converge; under out-of-sample stress they degrade gracefully---retaining far lower cost and SLO violations than both the exact solver and every baseline. We further evaluate rolling re-optimization on both synthetic demand volatility and a real Azure diurnal trace, finding that AGH's static plan is already robust enough that re-optimization mainly smooths peak-window cost rather than reducing mean cost.
\end{itemize}

\begin{figure*}[t]
    \centering
    \includegraphics[width=1\textwidth,height=0.14\textheight]{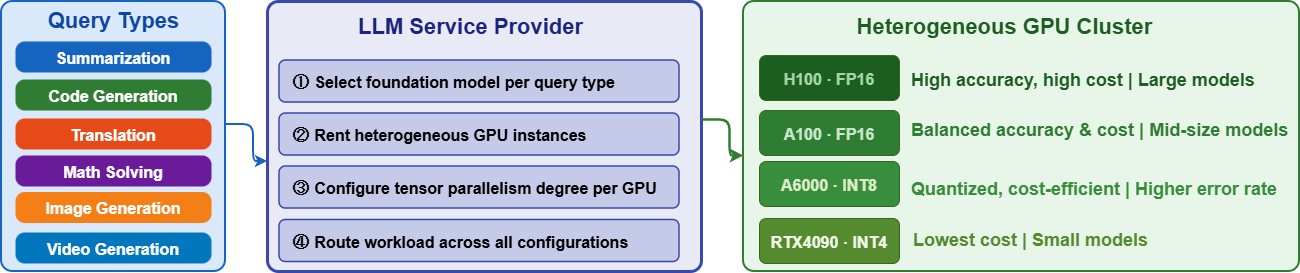}
    \caption{System model. Users submit queries that are classified into a finite set of types (summarization, code generation, translation, math solving, image generation, video generation). The SP hosts a catalog of foundation models on a set of heterogeneous GPU tiers; for each model--tier pair it chooses a tensor- and pipeline-parallelism configuration. Routing fractions split each query type's traffic across the deployed pairs. Latency, accuracy, memory, compute, storage, and budget constraints couple all four decisions.}
    \label{fig:LLM_model}
\end{figure*}

The remainder of the paper is organized as follows. Section~\ref{sec:related} reviews related work. Section~\ref{sec:model} presents the system model and MILP formulation. Section~\ref{sec:heuristics} describes the proposed heuristics and feasibility-preserving mechanisms. Section~\ref{sec:results} presents numerical results. Section~\ref{sec:conclusion} concludes the paper and discusses future directions.

\section{Related Work}
\label{sec:related}
This work is related to three lines of research: high-throughput LLM serving systems, optimization-based heterogeneous LLM placement, and 
heuristic approaches for cluster-level resource decisions. We discuss each direction and position our contribution relative to them.

\noindent\textbf{High-throughput LLM serving systems.}
This direction focuses on improving inference efficiency given a fixed deployment configuration. 
Representative techniques include continuous batching of in-flight requests for autoregressive decoding~\cite{yu2022orca}, paged-attention key--value cache management to reduce memory fragmentation~\cite{kwon2023vllm}, prefill--decode phase disaggregation that places the two phases on separate GPU pools to optimize goodput under per-phase service-level objectives~\cite{zhong2024distserve}, statistical multiplexing of model replicas to absorb burstiness under SLOs~\cite{li2023alpaserve}, and locality-aware request balancing across pre-deployed regional clusters~\cite{xia2025skylb}. These systems improve throughput and goodput at the execution layer but assume that model placement, GPU provisioning, and parallelism configurations are predetermined. \textit{Our work is complementary to this direction. Rather than optimizing inference execution for a fixed deployment, we address the upstream planning problem of jointly selecting models, provisioning heterogeneous GPU resources, configuring parallelism, and routing workload under system-level constraints.}


\noindent\textbf{Optimization-based LLM placement.}
Another line of work formulates heterogeneous LLM placement and routing as exact optimization problems~\cite{mei2025helix,jiang2025demystify}. These approaches use MILP or related formulations to jointly optimize deployment and workload assignment while enforcing resource and performance constraints. Such formulations can provide high-quality or optimal solutions for moderate problem sizes. However, the computational complexity of exact optimization grows rapidly with the size of the deployment space. As the numbers of query types, models, GPU tiers, and parallelism configurations increase, exact solvers become impractical for large-scale systems or dynamic environments requiring frequent re-optimization. In our experiments, the MILP solver exceeds a 600\,s time limit on production-scale instances, preventing rolling re-optimization under demand drift or GPU failures. \textit{Our work adopts a similar joint optimization perspective but focuses on scalable solution methods that preserve feasibility while enabling near-real-time operation.
}

\noindent\textbf{Heuristic and decomposed approaches.}
This direction applies fast, structured heuristics to one decision dimension at a time, treating the others as exogenous. Representative contributions reconfigure parallelism and GPU frequency on a homogeneous fleet to improve energy efficiency~\cite{stojkovic2024dynamollm}, adaptively select the best cloud-VM configuration for data-intensive workloads~\cite{alipourfard2017cherrypick}, 
decompose heterogeneous LLM serving into VM selection followed by routing~\cite{kim2025cost}, and share cluster resources across multiple co-served foundation models 
\cite{zhao2025seallm}. 
While these approaches achieve fast runtime, they typically optimize placement, provisioning, parallelism, and routing independently or sequentially. Hence, they do not explicitly preserve feasibility under tightly coupled system constraints such as per-GPU memory after sharding, two-phase latency under tensor/pipeline parallelism, and quantization-induced accuracy degradation. In LLM serving, generic cost-driven heuristics applied without feasibility-aware filtering can therefore produce allocations that violate memory or latency constraints. 


To evaluate this class of approaches, we benchmark against representatives of three heuristic families adapted to our joint formulation: an LP-relaxation with LP-warmstart greedy rounding (LPR), a decoupled VM-selection-then-routing scheme (DVR) inspired by Kim~et~al.~\cite{kim2025cost}, and a homogeneous-fleet provisioning scheme (HF) inspired by DynamoLLM~\cite{stojkovic2024dynamollm}.  
These baselines represent convex-relaxation, decomposed-allocation, and homogeneous-fleet planning approaches, respectively. Because existing methods target different decision spaces, our objective is not to reproduce each algorithm on its native formulation, but to evaluate representative heuristic families under the common deployment space and constraints considered in this paper. Empirical comparisons are presented in Section~\ref{sec:results}.


\noindent\textbf{Positioning of this work.}
This paper differs from prior work in three key aspects. First, parallelism configuration is treated as a decision variable jointly optimized with model placement, GPU provisioning, and workload routing rather than as a fixed system input. Second, feasibility under coupled memory, latency, accuracy, storage, and budget constraints is enforced directly within the heuristic loop. Third, the proposed heuristics operate within a runtime envelope suitable for rolling re-optimization at operational time scales. 
By combining these features, the proposed framework bridges the gap between exact optimization and fast heuristic allocation, enabling scalable and feasibility-preserving joint resource allocation for production-scale LLM inference systems.

\section{System Model and Problem Formulation}
\label{sec:model}

\subsection{System Model}
We consider a service provider (SP) that rents heterogeneous GPU instances from a public cloud platform to serve LLM inference workloads over a scheduling horizon $\Delta_T$, expressed in hours so that the per-hour GPU-rental and storage prices accrue over the horizon (e.g., the rental term scales as $\Delta_T p_k^c y_{j,k}$); we take $\Delta_T\!=\!24$\,h in our experiments. Let $i\in\mathcal{I}$ index query types, $j\in\mathcal{J}$ index foundation models, and $k\in\mathcal{K}$ index GPU resource tiers.
Table~\ref{tab:notation} summarizes the notation used in this paper. Fig.~\ref{fig:LLM_model} illustrates the system architecture. We first describe the four model components (queries, foundation models, GPU resource tiers, and parallelism) and then the two-phase delay model that ties them together; decision variables are formalized in Section~\ref{sec:opt}.

\noindent\textbf{\textit{Fixed deployment within an epoch.}} We treat each scheduling horizon $\Delta_T$ as one optimization epoch: the deployment variables (model placement, GPU provisioning, and TP/PP configuration) are held \emph{fixed} for the epoch, while routing adapts within the deployed configuration. This reflects operational practice: changing a deployment incurs non-negligible overhead, e.g., model loading, weight resharding, KV-cache invalidation, and serving-engine synchronization, so frequent reprovisioning within an epoch is not economical even when the planner itself is fast. Deployment is therefore re-optimized only at epoch boundaries, and the epoch length $\Delta_T$ is an operator-tunable parameter (in hours) that can be shortened or lengthened to match observed demand variability; Section~\ref{sec:rolling} studies a rolling variant that re-optimizes the deployment at a representative 5-minute \emph{cadence} within a fixed $\Delta_T\!=\!24$\,h horizon, rather than shortening $\Delta_T$ itself.

\begin{table}[t]
\caption{Summary of notation used throughout the paper.}
\label{tab:notation}
\centering
\footnotesize
\setlength{\tabcolsep}{4pt}
\renewcommand{\arraystretch}{1.05}
\begin{tabular}{@{}p{0.22\columnwidth} p{0.72\columnwidth}@{}}
\toprule
\textbf{Symbol} & \textbf{Description} \\
\midrule
\multicolumn{2}{@{}l}{\textit{Sets and indices}} \\
$i \in \mathcal{I}$ & query types and their index \\
$j \in \mathcal{J}$ & foundation models and their index \\
$k \in \mathcal{K}$ & GPU resource tiers and their index \\
$n \in \mathcal{N}_k$ & feasible TP degrees on tier $k$ \\
$m \in \mathcal{M}_k$ & feasible PP depths on tier $k$ \\
\midrule
\multicolumn{2}{@{}l}{\textit{Workload parameters}} \\
$\lambda_i$ & arrival rate of type $i$ (queries/h) \\
$h_i,\, f_i$ & average input / output token length \\
$r_i$ & total tokens per query, $r_i = h_i + f_i$ \\
$\theta_i$ & per-token storage footprint (KB/token) \\
\midrule
\multicolumn{2}{@{}l}{\textit{Foundation-model parameters}} \\
$B_j$ & weight footprint of model $j$ (GB) \\
$\beta_j$ & per-token KV-cache footprint (KB/token) \\
$\bar e_{i,j}^k$ & per-token error rate of $(j,k)$ on type $i$ \\
\midrule
\multicolumn{2}{@{}l}{\textit{GPU-tier parameters}} \\
$C_k^{\sf GPU}$ & per-GPU memory capacity (GB) \\
$P_k^{\sf GPU}$ & compute throughput (TFLOPs) \\
$p_k^c$ & hourly rental price (\$/h) \\
$\text{BW}_k$ & memory bandwidth (GB/s) \\
$\nu_k$ & quantization (latency) scale: 1.0, 0.5, 0.25 for FP16, INT8, INT4 \\
$\mu_k$ & quantization-induced error multiplier: 1, 1.15, 1.35 for FP16, INT8, INT4 \\
\midrule
\multicolumn{2}{@{}l}{\textit{Latency-model coefficients}} \\
$d_{i,j,k}^{\sf comp}$ & per-token compute delay at TP$=$1 (s) \\
$d_{i,j,k}^{\sf comm}$ & per-token inter-stage communication delay (s) \\
$\alpha_{i,j}^k$ & per-token compute cost (GFLOP/token) \\
$T_{i,j,k}^{\sf res}$ & KV-cache residency time per token \\
$\eta \in (0,1]$ & compute-utilization (PP-bubble) factor \\
$T_{\sf conv}$ & seconds-to-hours conversion, $3600$\,s/h \\
\midrule
\multicolumn{2}{@{}l}{\textit{SLOs, prices, and budgets}} \\
$\Delta_i,\, \epsilon_i$ & delay and error SLO of type $i$ \\
$\rho_i,\, \phi_i$ & delay and unmet-demand penalty rates \\
$p^s$ & per-GB-hour storage price \\
$\delta,\, C^s$ & global cost / storage caps \\
$\Delta_T$ & scheduling horizon (e.g., 24\,h) \\
$\zeta_i$ & per-type cap on the unmet-demand fraction \\
\midrule
\multicolumn{2}{@{}l}{\textit{Decision variables}} \\
$y_{j,k} \in \mathbb{Z}_+$ & GPUs allocated to pair $(j,k)$ \\
$q_{j,k} \in \{0,1\}$ & 1 if $(j,k)$ is deployed \\
$w_{j,k}^{n,m} \in \{0,1\}$ & joint TP/PP selector \\
$x_{i,j}^k \in [0,1]$ & fraction of type-$i$ traffic routed to $(j,k)$ \\
$z_{i,j}^k \in \{0,1\}$ & 1 if type $i$ is admitted on $(j,k)$ \\
$u_i \in [0,\zeta_i]$ & unserved fraction of type-$i$ demand \\
$v_{i,j}^{k,n,m}$ & McCormick auxiliary, $v = x_{i,j}^k\, w_{j,k}^{n,m}$ \\
\bottomrule
\end{tabular}
\end{table}

\noindent\textbf{\textit{1) Query types.}}
Users submit inference requests that the SP classifies into $I$ distinct query types, such as summarization, code generation, translation, math solving, image generation, and video generation. Query type $i$ is characterized by:
(a)~an arrival rate $\lambda_i$ (queries/hour), the time-average rate at which type-$i$ requests arrive over the scheduling horizon;
(b)~an average input length $h_i$ (tokens); and
(c)~an expected output length $f_i$ (tokens).
The total token count per query is $r_i \!=\! h_i + f_i$. 
The workload statistics used in our experiments are calibrated to the Azure LLM Inference Trace~\cite{azurellmtrace2025}; the calibration is detailed in Section~\ref{sec:results}.

\noindent\textbf{\textit{2) Foundation models.}}
The SP maintains a catalog of $J$ pre-trained foundation models with varying parameter sizes, ranging from lightweight (e.g., 1\,B parameters) to large-scale (e.g., 70\,B). Each model~$j$ is characterized by a weight footprint $B_j$ (GB) and a per-token key--value (KV) cache footprint $\beta_j$ (KB/token). Larger models generally yield higher output quality but incur higher per-GPU memory and per-token compute demands. The quality of serving query type $i$ using model $j$ on GPU tier $k$ is represented by a per-token error rate $\bar e_{i,j}^k$, which depends on both the model and the numerical precision used by the GPU tier. When type-$i$ traffic is served using model $j$ on GPU tier $k$, each query reserves KV-cache memory for a duration $T_{i,j,k}^{\sf res}$ (the \emph{residency time}), calibrated as the per-token decode duration in our experiments (Section~\ref{sec:results}).

\noindent\textbf{\textit{3) GPU resource tiers.}}
Inference jobs execute on heterogeneous GPU tiers indexed by $k$. Each tier combines a GPU hardware type with a numerical precision level (e.g., H100-FP16, A100-FP16, RTX\,4090-FP16, A6000-INT8, A6000-INT4). A tier captures both the physical device and the inference precision used on that device. Tier~$k$ is characterized by its per-GPU memory capacity $C_k^{\sf GPU}$ (GB), its compute throughput $P_k^{\sf GPU}$ (TFLOPs), the per-GPU hourly rental price $p_k^c$ (\$/h), and the memory bandwidth $\text{BW}_k$ (GB/s) that, together with the model weight footprint, governs the memory-bound decode time.

\noindent\textbf{\textit{4) Numerical precision and accuracy/cost trade-off.}}
Reduced-precision execution lowers memory and bandwidth usage but may increase inference error. We capture this trade-off using: 
a \emph{latency scaling factor} $\nu_k \in (0,1]$ that multiplies memory-bandwidth-bound decode time and the effective weight footprint, and an \emph{error multiplier} $\mu_k \geq 1$ that inflates the per-token error rate relative to the FP16 baseline. The effective error rate is modeled as:
\begin{equation}
\label{eq:error_calibration}
\bar e_{i,j}^k = \mu_k \cdot e_{i,j}^{\sf base},
\end{equation}
where $e_{i,j}^{\sf base}$ is the FP16 baseline error rate of model $j$ on query type $i$ and $\mu_k$ is calibrated to GPTQ-style post-training quantization~\cite{frantar2023gptq} ($\mu_k\!=\!1.0$ for FP16, $1.15$ for INT8, $1.35$ for INT4); the corresponding latency scale is $\nu_k\!=\!1.0, 0.5, 0.25$ for the same three precisions. The two scalars are independent calibrations that share the precision index but represent different physical effects: $\nu_k$ shrinks the bytes-per-weight that the memory subsystem must move, while $\mu_k$ captures the accuracy degradation introduced by post-training quantization. This precision-accuracy-cost trade-off is exactly what makes the deployment space non-trivial. For instance, a low-cost INT4 tier may violate strict accuracy SLOs for some query types, while an FP16 deployment that satisfies accuracy requirements may exceed the budget. Our proposed solution approaches in Section~\ref{sec:heuristics} use $\bar e_{i,j}^k$ to filter infeasible candidates before performing cost-based ranking.

\noindent\textbf{\textit{5) Parallelism configuration.}}
~Large models typically cannot fit on a single GPU and must therefore be distributed across multiple devices. We consider two complementary forms of parallelism: tensor parallelism (TP) and pipeline parallelism (PP).
In particular, \emph{TP} partitions model weights across $n$ co-located GPUs within a single pipeline stage. Each GPU then stores $1/n$ of the weights and performs $1/n$ of the per-token compute, accelerating the compute-bound prefill phase but introducing inter-GPU communication during autoregressive decoding.
\emph{PP} distributes the model layers across $m$ sequential pipeline stages, enabling larger models to fit across multiple GPU groups at the cost of pipeline bubbles (idle stages while a token traverses the pipeline).
For each deployed (model, tier) pair $(j,k)$, the SP selects a TP degree  $n \in \mathcal{N}_k$ and a PP depth  $m \in \mathcal{M}_k$ that jointly determine how many GPUs the pair consumes and how the model is sharded across them. The SP selects a TP degree $n \in \mathcal{N}_k$ from a hardware-dependent feasible set (e.g., $\{1,2,4,8\}$) and a PP depth $m \in \mathcal{M}_k$ from a system-wide feasible set (e.g., $\{1,2,4\}$). The binary variable $w_{j,k}^{n,m} \in \{0,1\}$ indicates whether $(j,k)$ uses the joint configuration $(\text{TP}=n,\,\text{PP}=m)$, and the deployment flag $q_{j,k} \in \{0,1\}$ records whether model~$j$ is active on tier~$k$. The total number of tier-$k$ GPUs allocated to 
pair $(j,k)$ is:
\begin{equation}
\label{eq:gpu_total}
    y_{j,k} = \!\!\sum_{(n,m) \in \mathcal{N}_k \times \mathcal{M}_k} \!\! n \cdot m \cdot w_{j,k}^{n,m}.
\end{equation}
These $y_{j,k}$ GPUs are organized as $\text{TP}_{j,k} = n$ tensor-parallel devices within each of $\text{PP}_{j,k} = m$ pipeline stages. This decomposition enters the model in two ways: TP governs per-stage memory and per-stage compute in the delay model~\eqref{eq:total_delay}; PP governs inter-stage communication overhead and induces a pipeline-bubble inefficiency that we capture as a multiplicative compute-utilization factor $\eta \in (0,1]$ in the compute constraint~\eqref{eq:compute} (set to $\eta\!=\!0.9$ in our experiments). Increasing TP reduces per-stage memory and compute cost, while increasing PP enables larger models to fit at the expense of pipeline inefficiency and communication overhead.

\noindent\textbf{\textit{6) Two-phase delay model.}} ~We decompose inference processing latency into two phases: (i) \emph{Prefill} phase that processes the input prompt and generates the first token (\textit{time-to-first-token, TTFT}); and (ii) \emph{Decode} phase that decodes and generates the remaining tokens autoregressively. Let $d_{i,j,k}^{\text{comp}}$ denote the per-token compute delay 
at TP$=$1 and $d_{i,j,k}^{\text{comm}}$ denote the per-token inter-stage communication delay under PP. Then the prefill latency is modeled as: 
\begin{align}
D_{i,j,k}^{\text{TTFT}} &= \frac{d_{i,j,k}^{\text{comp}}\, h_i}{\text{TP}_{j,k}},
\end{align}
and the decode latency is:
\begin{align}
D_{i,j,k}^{\text{Gen}} &= \Big( \frac{d_{i,j,k}^{\text{comp}}}{\text{TP}_{j,k}} + \text{PP}_{j,k}\, d_{i,j,k}^{\text{comm}}\Big)\, f_i.
\end{align}
Increasing TP reduces both terms because per-stage compute is sharded across more devices; increasing PP adds an additive inter-stage communication term that scales with the output length $f_i$. The expected end-to-end processing delay for query type~$i$ 
is:
\begin{align}
\label{eq:total_delay}
    D_i^{\text{proc}} = \sum_{j,k} x_{i,j}^k \bigg[ \frac{d_{i,j,k}^{\text{comp}} \cdot r_i}{\text{TP}_{j,k}} + \text{PP}_{j,k} \cdot d_{i,j,k}^{\text{comm}} \cdot f_i \bigg], ~~ \forall i,
\end{align}
where $x_{i,j}^k \in [0,1]$ is the fraction of type-$i$ workload routed to pair $(j,k)$ and $r_i \!=\! h_i + f_i$.

\noindent\textbf{\textit{7) MILP linearization.}}
Substituting $\text{TP}_{j,k} = n$ and $\text{PP}_{j,k} = m$ via the joint selector $w_{j,k}^{n,m}$ yields:
\begin{align}
\label{eq:proc_delay_milp}
    D_i^{\text{proc}} = \sum_{j,k} \sum_{(n,m)} x_{i,j}^k\, w_{j,k}^{n,m} \bigg[ \frac{d_{i,j,k}^{\text{comp}}\, r_i}{n} + m \cdot d_{i,j,k}^{\text{comm}}\, f_i \bigg], ~~ \forall i.
\end{align}
The only non-linearity is the bilinear terms $x_{i,j}^k \cdot w_{j,k}^{n,m}$ between a continuous $x_{i,j}^k \in [0,1]$ and a binary $w_{j,k}^{n,m} \in \{0,1\}$. We linearize each product using standard McCormick envelopes. Specifically, we introduce $v_{i,j}^{k,n,m} = x_{i,j}^k\, w_{j,k}^{n,m}$ with:
\begin{subequations}
\begin{align}
\label{eq:mccormick}
    & v_{i,j}^{k,n,m} \leq x_{i,j}^k, \quad v_{i,j}^{k,n,m} \leq w_{j,k}^{n,m}, \\
    & v_{i,j}^{k,n,m} \geq x_{i,j}^k + w_{j,k}^{n,m} - 1, \quad v_{i,j}^{k,n,m} \geq 0.
\end{align}
\end{subequations}
Note that no trilinear term arises because TP and PP are jointly encoded by the single binary variable $w_{j,k}^{n,m}$. Thus, a single McCormick linearization layer suffices to represent all delay terms exactly. We define the per-configuration constant $D_{i,j}^k(n,m) \!=\! d_{i,j,k}^{\text{comp}} r_i / n + m\, d_{i,j,k}^{\text{comm}}\, f_i$, so that $D_i^{\text{proc}} = \sum_{j,k,(n,m)} v_{i,j}^{k,n,m}\, D_{i,j}^k(n,m)$ is linear in the auxiliary variables.

\subsection{Optimization Problem}
\label{sec:opt}
The SP jointly optimizes model deployment, heterogeneous GPU provisioning, tensor/pipeline parallelism configuration, workload routing, and unmet demand. The resulting problem is formulated as an MILP that minimizes total operational cost while satisfying latency, accuracy, memory, compute, storage, and budget constraints.

\noindent\textbf{Decision variables.} 
There are four groups of variables:
\begin{itemize}\setlength{\itemsep}{1pt}
\item \emph{Provisioning and deployment.} The integer GPU count $y_{j,k} \in \mathbb{Z}_+$ allocated to each pair $(j,k)$, and the binary deployment indicator $q_{j,k} \in \{0,1\}$ indicating whether $(j,k)$ is active.
\item \emph{Parallelism.} The joint TP/PP selector $w_{j,k}^{n,m} \in \{0,1\}$, with $\sum_{n,m} w_{j,k}^{n,m} = q_{j,k}$ enforcing exactly one configuration per active pair.
\item \emph{Routing and placement.} 
$x_{i,j}^k \in [0,1]$ represents the fraction of query type $i$ routed to $(j,k)$, and $z_{i,j}^k \in \{0,1\}$ indicates whether type $i$ is admitted on $(j,k)$.
\item \emph{Unserved demand.} The continuous slack variable $u_i \in [0, \zeta_i]$ denotes the fraction of type-$i$ demand that remains unserved (with a per-type cap $\zeta_i$).
\end{itemize}

The deterministic placement problem $\mathcal{P}_{\sf DM}$ minimizes total operational cost over the scheduling horizon:
\begin{subequations}
\label{eq:DM}
\begin{align}
    & \mathcal{P}_{\sf DM}\!:  \min_{\bx,\by,\bz,\bu,\bw} ~~
    \underbrace{\Delta_T \!\sum_{j,k} p_k^c \, y_{j,k}}_{\text{(i) GPU rental}}
    + \underbrace{\Delta_T \!\sum_{i,j,k} p^s B_j \, z_{i,j}^k}_{\text{(ii) model storage}}  \notag\\
    & + \underbrace{\Delta_T \!\sum_{i,j,k} p^s \theta_i r_i \lambda_i \, x_{i,j}^k}_{\text{(iii) data storage}} + \underbrace{\sum_{i} \rho_i \, D_i^{\text{proc}}}_{\text{(iv) delay penalty}}  + \underbrace{\Delta_T \!\sum_{i} \phi_i \, u_i}_{\text{(v) unmet penalty}} \label{eq:DM_obj}\\
     & \text{s.t.} ~ \sum_{j,k} x_{i,j}^k + u_i = 1, ~~ \forall i \label{eq:demand}\\
    &  \Delta_{T}  \sum_{j,k} p^{\sf c}_k y_{j,k} + \Delta_{T} \sum_{i,j,k} p^s \big( B_j z_{i,j}^{k} + \theta_i (h_i + f_i) \lambda_i x_{i,j}^{k} \big) \leq \delta \label{constr:budget} \\
    & \sum_{(n,m) \in \mathcal{N}_k \times \mathcal{M}_k} \!\! w_{j,k}^{n,m} = q_{j,k},~ \forall j,k \label{eq:tp_select}\\
    & y_{j,k} = \!\!\sum_{(n,m) \in \mathcal{N}_k \times \mathcal{M}_k} \!\! n \cdot m \cdot w_{j,k}^{n,m}, ~~ \forall j,k \label{eq:y_def}\\
    & \sum_{n,m} \frac{B_j}{nm} w_{j,k}^{n,m} \nonumber \\
    &\quad + \sum_{n,m} \frac{\beta_j}{nm} w_{j,k}^{n,m} \cdot \sum_i r_i T_{i,j,k}^{\sf res} x_{i,j}^k \leq C_k^{\sf GPU} q_{j,k}, ~\forall j,k \label{eq:mem}\\
    & \sum_{i} \alpha_{i,j}^k \bigg(\frac{r_i \, \lambda_{i} }{10^3} \bigg)  x_{i,j}^{k} \leq \eta \, T_{\mathrm{conv}} \, P_{k}^{\sf GPU} \, y_{j,k}, ~~ \forall j,k \label{eq:compute}\\
    & \sum_{j,k} B_j \, z_{i,j}^k + \theta_i r_i \lambda_i x_{i,j}^k \leq C^s \label{eq:storage}\\
    & D_i^{\text{proc}} \leq \Delta_i, ~ \forall i \label{eq:delay_constr}\\
    & \sum_{j,k} \bar{e}_{i,j}^k \, x_{i,j}^k \leq \epsilon_i, ~ \forall i \label{eq:error_constr}\\
    & 0 \leq x_{i,j}^k \leq z_{i,j}^k \leq q_{j,k},
      ~ \forall i,j,k \label{eq:link}
\end{align}
\end{subequations}

\noindent\textbf{Objective.} The five cost components in~\eqref{eq:DM_obj} capture the SP's operational cost over the horizon~$\Delta_T$. The first three terms represent resource provisioning costs, including GPU rental, model-weight storage, and workload-associated data storage. The remaining two terms capture SLO-related penalties. 
The running costs comprise GPU rental ($\Delta_T p_k^c y_{j,k}$, the hourly price of every allocated GPU), model-weight storage ($\Delta_T p^s B_j z_{i,j}^k$, paid per (model, tier) pair that admits type $i$ at the per-GB-hour rate $p^s$), and data storage ($\Delta_T p^s \theta_i r_i \lambda_i x_{i,j}^k$, where $\theta_i$ is the per-token footprint in KB/token), each scaled to the scheduling horizon. The remaining two terms shape SLO behavior at the optimum: $\rho_i D_i^{\text{proc}}$ is a soft delay surrogate (a per-type rate of \$/ms/query for delay-induced revenue loss), with the hard delay bound enforced separately by~\eqref{eq:delay_constr}; and $\Delta_T \phi_i u_i$ penalizes dropped traffic over the horizon, where $\phi_i$ is the per-hour cost of leaving type-$i$ demand fully unserved (placed on the same horizon scale $\Delta_T$ as the rental and storage terms), chosen much larger than per-GPU rental so that admission is preferred whenever a feasible placement exists.

\noindent\textbf{Constraints.} The nine constraint groups fall into four operational categories. \emph{Demand and budget.} Equation~\eqref{eq:demand} ensures that every unit of type-$i$ demand is either routed or recorded as unmet, and~\eqref{constr:budget} caps the horizon-aggregated GPU rental, weight storage, and per-token storage at the SP-wide budget $\delta$. \emph{Parallelism configuration.} Equations~\eqref{eq:tp_select}--\eqref{eq:y_def} require each active pair $(j,k)$ to use exactly one (TP, PP) joint configuration and to provision $y_{j,k}$ GPUs equal to the chosen $\text{TP}\!\times\!\text{PP}$ product, eliminating any inconsistency between the binary selector $w_{j,k}^{n,m}$ and the integer provisioning variable. \emph{Resource capacity.} Three bounds enforce hardware budgets: the per-GPU memory bound~\eqref{eq:mem} requires the weight shard $B_j/(nm)$ plus the KV-cache footprint $\beta_j/(nm)$ aggregated over served tokens (with per-token residency $T_{i,j,k}^{\sf res}$) to fit in $C_k^{\sf GPU}$, where TP shards weights by factor $1/n$ and PP shards layers by factor $1/m$; the compute-throughput bound~\eqref{eq:compute} requires aggregate per-token FLOPs routed to $(j,k)$ to fit within the pair's hourly capacity, where $\alpha_{i,j}^k$ is the per-token compute cost (GFLOP/token), the $10^3$ factor aligns token units with the TFLOP-denominated $P_k^{\sf GPU}$, $T_{\mathrm{conv}}\!=\!3600$\,s/h converts seconds to hours, and $\eta\!\leq\!1$ captures pipeline-bubble inefficiency; and~\eqref{eq:storage} caps the total weight footprint plus per-token data on $(j,k)$ at the SP-wide storage budget $C^s$. \emph{SLO enforcement and routing consistency.} The hard delay SLO~\eqref{eq:delay_constr} enforces $D_i^{\text{proc}}\!\leq\!\Delta_i$ per query type and is precisely what makes the bilinear term $x_{i,j}^k\, w_{j,k}^{n,m}$ visible to the solver, motivating the McCormick auxiliaries $v_{i,j}^{k,n,m}$; the error SLO~\eqref{eq:error_constr} caps the traffic-weighted per-token error rate at $\epsilon_i$, which is the mechanism that rules out routing strict-accuracy queries to deeply quantized tiers; finally,~\eqref{eq:link} chains $0 \!\le\! x_{i,j}^k \!\le\! z_{i,j}^k \!\le\! q_{j,k}$ to forbid ``ghost'' routing to undeployed placements.

\noindent\textbf{Variable counts and motivation for heuristics.}
$\mathcal{P}_{\sf DM}$ has $O(IJK)$ continuous routing variables, $O(IJK)$ binary placement indicators, $O(JK)$ binary deployment indicators, $O(JK|\mathcal{N}||\mathcal{M}|)$ binary configuration selectors, and $O(IJK|\mathcal{N}||\mathcal{M}|)$ McCormick auxiliary variables. With $|\mathcal{N}|=4$, $|\mathcal{M}|=3$, and $(I,J,K)=(20,20,20)$, this corresponds to approximately $1.3\times10^4$ binary variables and $10^5$ auxiliary variables, motivating the scalable heuristics developed in Section~\ref{sec:heuristics}.

\vspace{-0.3cm}
\section{Solution Approach}
\label{sec:heuristics}

The MILP $\mathcal{P}_{\sf DM}$ can be solved exactly for moderate instances, but its runtime grows exponentially with the problem size and exceeds a 600\,s time limit on the largest instance in our experiments. Moreover, the SP may need to re-solve the allocation problem at the cadence of demand drift (e.g., every 5 minutes in our rolling-horizon study), so a fast heuristic that produces feasible, near-optimal solutions is operationally essential.

The challenge specific to this problem is that the constraints are \emph{tightly coupled}: per-GPU memory~\eqref{eq:mem} limits which TP degrees fit a given model on a given tier, the TP/PP choice directly enters the delay SLO~\eqref{eq:delay_constr}, the error SLO~\eqref{eq:error_constr} couples to routing through $\bar e_{i,j}^k$, and the global budget~\eqref{constr:budget} caps the total set of activated GPUs. A generic greedy that ranks candidates by raw cost ignores these dependencies and routinely yields infeasible solutions, as our ablation in Section~\ref{sec:results} confirms. The three mechanisms below close the gap; they are shared by GH and AGH.

\noindent\textbf{Running state shared by all mechanisms.} Both algorithms maintain the following auxiliary state during construction (we use these symbols throughout the rest of this section):
\begin{itemize}\setlength{\itemsep}{1pt}
\item $\mathcal{I}^{\sf unc}$ = the \emph{uncovered set}: query types $i \in \mathcal{I}$ that have no admitted placement so far. Initialized to $\mathcal{I}$, this set drives Phase~1 (coverage pre-allocation) and shrinks each time a new $(j^*,k^*)$ pair is activated.
\item $\tilde r_i \in [0,1]$ = the \emph{remaining unserved fraction} of demand for type $i$. Initialized to $1$ and decremented as $x_{i,j}^k$ is committed.
\item $E_i^{\sf used} = \sum_{j,k} \bar e_{i,j}^k\, x_{i,j}^k$ = the \emph{cumulative error} accrued by type $i$ from previously committed routing. The remaining error budget is $\epsilon_i - E_i^{\sf used}$.
\item $D_i^{\sf used} = \sum_{j,k}\, x_{i,j}^k\, D_{i,j}^k(n^*,m^*)$ = the \emph{cumulative delay} similarly. The remaining delay budget is $\Delta_i - D_i^{\sf used}$.
\end{itemize}

\subsection{Three Constraint-Aware Mechanisms}

\subsubsection{M1: TP-Aware Feasibility Selection}
For every candidate triple $(i,j,k)$, the algorithm selects the cheapest (TP, PP) configuration that \emph{simultaneously} satisfies the per-GPU memory budget and the delay SLO:
\begin{align}
\label{eq:tp_select_mech}
\begin{split}
    (n^*\!,m^*)(i,j,k) &= \argmin_{(n,m) \in \mathcal{N}_k \times \mathcal{M}_k}\!\bigg\{nm \!: \\
    &\qquad \frac{B_j}{nm} \leq C_k^{\sf GPU},~ D_{i,j}^k(n,m) \leq \Delta_i \bigg\}.
\end{split}
\end{align}
If no $(n,m)$ satisfies both inequalities, the candidate $(i,j,k)$ is \emph{discarded entirely}: it cannot serve type $i$ on tier $k$ no matter how attractive its cost. This is the mechanism that prevents the failure modes of generic cost-only greedies: a 70\,B model on a 24\,GB tier under TP$\!=\!$1, or a small model on a slow tier where TTFT alone exceeds $\Delta_i$. We will see in Section~\ref{sec:results} that disabling M1 produces infeasible solutions (memory or delay violations), not merely worse ones.

\subsubsection{M2: Cost-Per-Effective-Coverage Ranking}
Once M1 has fixed an admissible $(\hat n, \hat m)\!=\!(n^*\!,m^*)(i,j,k)$, the algorithm ranks candidates not by raw cost but by cost per unit of \emph{effective coverage}, namely the maximum traffic fraction that the candidate can absorb without violating the remaining error or delay budget. The marginal cost of placing query type~$i$ on $(j,k)$ at configuration $(\hat n, \hat m)$ is:
\begin{align}
\label{eq:marginal_cost}
    \!\! c_{i,j}^k \!=\! \Delta_T \big[ p_k^c (\hat{n}\hat{m} \!-\! y_{j,k})^+ \!+\! p^s(B_j \!+\! \theta_i r_i \lambda_i) \big] + \rho_i D_{i,j}^k(\hat{n},\hat{m}),
\end{align}
where $(\hat{n}\hat{m} - y_{j,k})^+$ is the \emph{incremental} GPU cost (zero if $(j,k)$ is already active with at least $\hat n \hat m$ GPUs, see M3 below). The effective coverage is the maximum traffic fraction that fits the remaining error and delay budgets:
\begin{equation}
\label{eq:eff_coverage}
    \bar{x}_{i,j}^k = \min\!\bigg(\tilde{r}_i,~ \frac{\epsilon_i - E_i^{\sf used}}{\bar{e}_{i,j}^k},~ \frac{\Delta_i - D_i^{\sf used}}{D_{i,j}^k(\hat{n},\hat{m})}\bigg).
\end{equation}
Candidates are sorted lexicographically by $(\pi, \kappa)$, where the binary tie-breaker $\pi \!=\! \mathbf{1}[\bar{x}_{i,j}^k \!<\! \tilde{r}_i]$ prioritizes \emph{full-coverage} candidates (those that can absorb all of the remaining demand at once, $\pi\!=\!0$), and $\kappa \!=\! c_{i,j}^k / \bar{x}_{i,j}^k$ is the unit cost. The effect is that the heuristic refuses to spend its budget on cheap-but-narrow candidates whose error or delay slack is too small to clear the remaining demand.

\subsubsection{M3: TP Upgrade on Active GPUs}
When query type~$i$ is to be routed to an \emph{already-active} pair $(j,k)$ that currently holds $y_{j,k}$ GPUs, but the existing configuration's delay exceeds $\Delta_i$, the algorithm seeks a higher-parallelism configuration on the same pair before considering activating a fresh pair:
\begin{multline}
\label{eq:tp_upgrade}
    (\hat{n},\hat{m}) = \argmin_{(n,m) \in \mathcal{N}_k\times\mathcal{M}_k}\!\big\{ nm > y_{j,k} : \\
    D_{i,j}^k(n,m) \leq \Delta_i,~ \text{budget allows}\big\}.
\end{multline}
The pair pays only the incremental cost $(\hat n \hat m - y_{j,k})\,p_k^c \Delta_T$ rather than the full activation cost of a fresh pair: the model weights are already loaded, so the incremental GPUs simply share the existing TP/PP group. Like M1, M3 is a feasibility prerequisite: without it, queries that arrive late in the construction order may have no admissible target left within the budget.

\begin{algorithm}[t]
\caption{Greedy Heuristic (GH)}
\label{alg:GH}
\begin{algorithmic}[1]
\REQUIRE Sets $\mathcal{I},\mathcal{J},\mathcal{K},\mathcal{N}_k,\mathcal{M}_k$; coefficients $d_{i,j,k}^{\sf comp},d_{i,j,k}^{\sf comm},\bar e_{i,j}^k$; SLOs $\Delta_i,\epsilon_i$; budget $\delta$.
\ENSURE Allocation $(\bx, \by, \bz, \bu)$

\STATE Initialize $x_{i,j}^k \!\leftarrow\! 0$, $y_{j,k} \!\leftarrow\! 0$, $u_i \!\leftarrow\! 1$, $\tilde r_i \!\leftarrow\! 1$, $E_i^{\sf used}\!\leftarrow\!0$, $D_i^{\sf used}\!\leftarrow\!0$, $\mathcal{I}^{\sf unc} \!\leftarrow\! \mathcal{I}$ \hfill (uncovered set)
\STATE \textit{// Phase 1: Coverage pre-allocation}
\WHILE{$\mathcal{I}^{\sf unc} \neq \emptyset$ \AND budget $< \beta \cdot \delta$}
    \STATE Compute $\mathcal{F}_{j,k}$ via~\eqref{eq:coverage_set} and $\text{Cost}(j,k)$ via~\eqref{eq:activation_cost}, $\forall j,k$
    \STATE Activate $(j^*\!,k^*) \!=\! \argmax_{j,k} |\mathcal{F}_{j,k}|/\text{Cost}(j,k)$; update $\mathcal{I}^{\sf unc}$, budget, $\by$
\ENDWHILE
\STATE \textit{// Phase 2: Sequential allocation}
\FOR{each query $i$ sorted by $\lambda_i$ descending}
    \FOR{each candidate $(j,k)$}
        \STATE \textit{Step 1:} Determine $(\hat{n},\hat{m})$ via M1~\eqref{eq:tp_select_mech} or M3~\eqref{eq:tp_upgrade}
        \STATE \textit{Step 2:} Compute $\bar{x}_{i,j}^k$ via~\eqref{eq:eff_coverage}; discard if $\leq 0$
        \STATE \textit{Step 3:} Compute $c_{i,j}^k$ via~\eqref{eq:marginal_cost}; record $(\pi, \kappa)$
    \ENDFOR
    \STATE Sort candidates by $(\pi, \kappa)$ ascending
    \FOR{each $(j,k)$ in sorted order while $u_i > 0$}
        \STATE \textit{Step 4:} Verify~\eqref{eq:mem}--\eqref{eq:storage} and budget $\delta$
        \IF{all constraints satisfied}
            \STATE $x_{i,j}^k \!\leftarrow\! \min(u_i, \bar{x}_{i,j}^k)$; ~$u_i \!\leftarrow\! u_i - x_{i,j}^k$
            \STATE Update $E_i^{\sf used}$, $D_i^{\sf used}$, $\by$, budget
        \ENDIF
    \ENDFOR
\ENDFOR
\RETURN $(\bx, \by, \bz, \bu)$
\end{algorithmic}
\end{algorithm}

\subsection{Greedy Heuristic (GH)}
GH performs a single-pass allocation in two phases, invoking \textbf{M1}--\textbf{M3} throughout to ensure feasibility at every step.

\textit{\textbf{Phase~1: Coverage pre-allocation}} (lines~2--5) operates as a greedy set-cover that activates one pair $(j,k)$ at a time, choosing each activation to maximize the number of currently uncovered query types it admits per dollar of rental. We characterize each candidate pair through two coupled quantities. First, the \emph{coverage set} of $(j,k)$ collects the currently uncovered types that the pair could feasibly serve, namely those for which mechanism M1 returns a feasible (TP, PP) configuration $(n^*, m^*)(i,j,k)$ and whose per-token error rate is within the SLO:
\begin{equation}
\label{eq:coverage_set}
    \mathcal{F}_{j,k} = \big\{ i \in \mathcal{I}^{\sf unc} : (n^*\!,m^*)(i,j,k) \text{ exists via M1},\; \bar{e}_{i,j}^k \leq \epsilon_i \big\}.
\end{equation}
Second, because Phase~1 commits to a single configuration on $(j,k)$ that must accommodate \emph{every} admitted type, the GPUs allocated to the pair must be enough for the most demanding member of its coverage set. The corresponding \emph{activation cost} is the horizon-aggregated rental of that group:
\begin{equation}
\label{eq:activation_cost}
    \text{Cost}(j,k) = \Delta_T \, p_k^c \, \max_{i \in \mathcal{F}_{j,k}} n^*(i,j,k) \!\cdot\! m^*(i,j,k),
\end{equation}
where $\max_{i \in \mathcal{F}_{j,k}} n^*(i,j,k)\, m^*(i,j,k)$ is the smallest GPU count under which the chosen (TP, PP) shard simultaneously satisfies all types in $\mathcal{F}_{j,k}$. The algorithm then greedily selects $(j^*\!,k^*) = \argmax_{j,k} |\mathcal{F}_{j,k}|/\text{Cost}(j,k)$, i.e., the pair that covers the most uncovered types per unit cost, and repeats until either the uncovered set is empty or the cumulative rental reaches the Phase~1 budget cap $\beta\delta$ (with $\beta\!=\!0.8$).

\textit{\textbf{Phase~2: Sequential allocation}} (lines~6--20) processes queries in descending $\lambda_i$ order. For each query~$i$ and candidate $(j,k)$: (1)~determine $(\hat{n},\hat{m})$ via M1~\eqref{eq:tp_select_mech} or M3~\eqref{eq:tp_upgrade}, discarding infeasible candidates; (2)~compute effective coverage $\bar{x}_{i,j}^k$ via~\eqref{eq:eff_coverage}; (3)~rank by $(\pi, \kappa)$ where $\pi\!=\!\mathbf{1}[\bar{x}_{i,j}^k \!<\! \tilde{r}_i]$ prioritizes full-coverage and $\kappa\!=\!c_{i,j}^k/\bar{x}_{i,j}^k$ is unit cost via~\eqref{eq:marginal_cost}; and (4)~verify constraints~\eqref{eq:mem}--\eqref{eq:storage} and budget~$\delta$ before committing $x_{i,j}^k \!=\! \min(u_i, \bar{x}_{i,j}^k)$.

\subsection{Adaptive Greedy Heuristic (AGH)}

GH is efficient and feasibility-safe, but its single-pass structure has three structural limitations: (i)~solution quality depends on the order in which query types are processed in Phase~2; (ii)~once a workload fraction is committed, it cannot be revised even if a better candidate appears later in the construction; and (iii)~GPUs activated early in Phase~1 to cover narrow query types may remain underutilized after Phase~2 redistributes traffic. AGH (Algorithm~\ref{alg:AGH}) targets these three weaknesses with three corresponding enhancements while still calling GH's M1--M3 inside each construction step:

\begin{itemize}\setlength{\itemsep}{1pt}
\item \textbf{Multi-start construction} (Algorithm~\ref{alg:AGH}, lines~2--5): generates eight deterministic orderings of query types (ascending and descending each for $\lambda_i$, $\phi_i$, weight footprint $B_j$ as it appears for that type, and error tightness $\epsilon_i$) plus $R$ random permutations, runs GH for each, and keeps the best feasible solution. This addresses (i): a single ordering can trap GH at a poor solution that a different ordering avoids.
\item \textbf{Relocate local search} (lines~6--9): up to $L\!=\!3$ passes that, for each currently active assignment $(i,j,k)$ with $x_{i,j}^k\!>\!0$, attempt to move it to an alternative pair $(j'\!,k')$. A move is accepted if it remains feasible under all constraints of $\mathcal{P}_{\sf DM}$ \emph{and} strictly improves the objective. This addresses (ii) by allowing post-construction revision.
\item \textbf{Consolidation} (lines~10--12): scans active pairs $(j,k)$ in ascending order of GPU load, attempts to redistribute their traffic to other active pairs, and deactivates a pair if all of its traffic can be reabsorbed feasibly with strict cost improvement. This addresses (iii) and shrinks the GPU rental term in the objective.
\end{itemize}


\begin{algorithm}[t]
\caption{Adaptive Greedy Heuristic (AGH)}
\label{alg:AGH}
\begin{algorithmic}[1]
\REQUIRE $R$ random starts, max local search iterations $L$
\ENSURE Best allocation $(\bx^*, \by^*, \bz^*, \bu^*)$

\STATE $\text{best\_obj} \leftarrow \infty$
\STATE Generate orderings $\Omega \!=\! \{\omega_1, \ldots, \omega_8\} \cup \{R \text{ random permutations}\}$
\hfill [$R$ adaptive, see Remark~\ref{rem:agh_complexity}]
\FOR{each ordering $\omega \in \Omega$}
    \STATE $(\bx, \by, \bz, \bu) \leftarrow \textsc{GH\text{-}Construct}(\omega)$ \hfill [M1, M2, M3]
    \STATE \textit{// Local Search: relocate}
    \FOR{iter $= 1, \ldots, L$}
        \FOR{each $(i,j,k)$ with $x_{i,j}^k > 0$}
            \STATE Try move to $(j'\!,k')$; accept if feasible \& cost-improving
        \ENDFOR
    \ENDFOR
    \STATE \textit{// Local Search: consolidate}
    \FOR{each active $(j,k)$ in ascending order of load}
        \STATE Redistribute queries; deactivate $(j,k)$ if feasible \& improving
    \ENDFOR
    \IF{objective of $(\bx,\by,\bz,\bu) < \text{best\_obj}$}
        \STATE $(\bx^*\!, \by^*\!, \bz^*\!, \bu^*) \!\leftarrow\! (\bx, \by, \bz, \bu)$; ~update best\_obj
    \ENDIF
\ENDFOR
\RETURN $(\bx^*, \by^*, \bz^*, \bu^*)$
\end{algorithmic}
\end{algorithm}

\subsection{Complexity Analysis}
\label{complexity}
\begin{remark}[GH complexity]
\label{rem:gh_complexity}
GH runs in $O(I^2 J K + I J K \log(JK))$. The first term comes from the Phase~1 set-cover, which performs at most $I$ activations and re-evaluates $\mathcal{F}_{j,k}$ across $JK$ candidate pairs at each activation, giving $O(IJK)$ work per activation. The second term comes from Phase~2: each of the $I$ query types ranks $JK$ candidates in $O(JK \log(JK))$ time and commits feasible assignments in linear time. The constant factors are small because mechanisms M1--M3 are constant-work-per-candidate filters.
\end{remark}

\begin{remark}[AGH complexity]
\label{rem:agh_complexity}
AGH executes $(8 + R)$ independent constructions: 8 deterministic orderings (ascending and descending $\lambda_i$, $\phi_i$, storage footprint, and error tightness $\epsilon_i$) plus $R$ random permutations. Each construction calls GH and is followed by up to $L$ relocate passes, each with worst-case cost $O(L\, I^2 J^2 K^2)$, while the relocate cost dominates the consolidation pass. The overall complexity is therefore
\[
O\big( (8+R)\,[\,I^2 J K + I J K \log(JK) + L\, I^2 J^2 K^2\,] \big).
\]
The random-start count $R$ adapts to the problem scale $N = IJK$: $R\!=\!3$ for $N \!>\! 5000$, $R\!=\!5$ for $N \!>\! 2000$, $R\!=\!10$ for $N \!>\! 500$, and $R\!=\!20$ otherwise. Construction terminates early after five consecutive non-improving orderings, and we set $L\!=\!3$ throughout.
\end{remark}

\noindent \textbf{\textit{Fixed deployment within an epoch.}} In our framework, each scheduling horizon $\Delta_T$ is treated as a single optimization epoch. Deployment decisions, including model placement, GPU provisioning, and TP/PP configuration, remain \emph{fixed} throughout the epoch, while routing decisions adapt within the deployed configuration.
Although our heuristics execute within a few seconds, changing deployments in practice incurs non-negligible overhead, including model loading, weight resharding, KV-cache invalidation, and serving-engine synchronization. Consequently, faster re-optimization does not directly translate into continuous reprovisioning. Deployments are therefore updated only at epoch boundaries, while routing dynamically adapts within the active deployment. The epoch length is operator-configurable and can be adjusted according to observed demand variability. Section~\ref{sec:rolling} studies a rolling-epoch variant with a representative 5-minute setting.

\section{Numerical Results}
\label{sec:results}

\subsection{Simulation Setup}
\label{sec:setup}

\noindent\textbf{Lattice and parameter ranges.}
The base instance has $I\!=\!6$ query types (Summarization, Code Generation, Translation, Math Solving, Image Generation, Video Generation), $J\!=\!6$ Llama-3.x foundation models~\cite{dubey2024llama3} with weight footprints $B_j$ ranging from 2\,GB (1\,B parameters) to 140\,GB (70\,B parameters) and per-token KV-cache footprints $\beta_j$ in the range 31--305\,KB/token, and $K\!=\!10$ GPU tiers drawn from the cross product of the hardware set $\{$A6000, RTX\,4090, A100-40\,GB, H100-80\,GB$\}$ and the precision set $\{$FP16, INT8, INT4$\}$, excluding A100-INT4 and H100-INT4 (not deployed in our setup, as data-center quantization on those tiers is rarely offered at production rates).\footnote{Per-GPU memory ranges 24--80\,GB; memory bandwidth 768--3350\,GB/s; compute throughput 40.7--1484\,TFLOPs; values from NVIDIA datasheets.} For simplicity our experiments take $\mathcal{N}_k\!=\!\mathcal{N}\!=\!\{1,2,4,8\}$ and $\mathcal{M}_k\!=\!\mathcal{M}\!=\!\{1,2,4\}$ identically across $k$; with the joint selector $w_{j,k}^{n,m}$ this gives 12 candidate configurations per pair and bilinear delay terms that require only a single McCormick layer (Section~\ref{sec:opt}). For runtime-scaling experiments we expand $(I,J,K)$ up to $(20,20,20)$.

\noindent\textbf{Workload calibration to the Azure LLM Inference Trace.}
We calibrate arrival rates and token-length distributions to the public Azure LLM Inference Trace~\cite{azurellmtrace2025}, which contains anonymized OpenAI-style request logs collected from a production Azure cluster, including timestamps, input-token counts, and generation lengths. Specifically, (a) we extract the \texttt{ContextTokens} and \texttt{GeneratedTokens} fields and the request timestamp; (b) since the trace does not include explicit query-type labels, we map requests into our six query types by jointly bucketing on input length, output length, and output/input ratio using thresholds informed by Splitwise~\cite{patel2024splitwise} (e.g., short input and long output $\Rightarrow$ Code/Math; long input and short output $\Rightarrow$ Summarization), while image- and video-generation subsets are calibrated separately because the public trace is text-only; (c) the per-type arrival rate $\lambda_i$ is set to the empirical hourly rate within each bucket and anchored to the orders of magnitude reported by Splitwise, ranging from 1{,}000--3{,}000\,queries/h for Video Generation to 18{,}000--25{,}000\,queries/h for Summarization; and (d) $h_i$ and $f_i$ are taken as the bucket means. This preserves the heavy-tailed token-length distribution within each class while making the $I\!=\!6$ class structure explicit. Image and Video Generation arrivals, which are absent from the text-only trace, are calibrated using the rates reported in~\cite{patel2024splitwise} and placed at the lower end of the lattice (1{,}000--3{,}000\,queries/h); this is the only non-trace-derived component of the workload.

\noindent\textbf{SLOs, prices, and penalties.}
Delay SLOs $\Delta_i$ range from 1.5\,s for the most latency-sensitive types to 25\,s for video generation~\cite{stojkovic2024dynamollm}; per-token error tolerances satisfy $\epsilon_i \in [2\%,8\%]$. GPU rental prices satisfy $p_k^c \in [\$0.35,\$2.50]$/h and follow market rates compiled in~\cite{wilkinsoffline}; the horizon budget is $\delta\!=\!\$100$ and $\Delta_T\!=\!24$\,h, so hourly rental and storage costs accrue over the horizon, while the rolling study in Section~\ref{sec:rolling} re-optimizes every 5\,min within this horizon. The SP-wide storage cap is $C^s\!=\!1{,}000$\,GB and the Phase-1 budget fraction is $\beta\!=\!0.8$. Storage price is $p^s \sim \mathcal{U}[0.0005,0.001]$\,\$/GB-h. Delay penalties are task-tiered: $\rho_i \in [0.0001,0.0003]$\,\$/ms/query for text tasks, $[0.0005,0.0008]$ for Math Solving, and $[0.0005,0.001]$ for Image/Video Generation. The unmet-demand penalty $\phi_i$ is \$500--\$750 per hour of fully unserved demand for text tasks and \$1{,}000--\$1{,}500 for media-generation tasks, so leaving a type entirely unmet over the horizon incurs $\phi_i\Delta_T$ on the same scale as the rental and storage terms. Per-token storage footprints $\theta_i$~\cite{kwon2023vllm} are 10--14\,KB/token for text, 40--60\,KB/token for image, and 80--120\,KB/token for video.

\noindent\textbf{Latency model parameters.}
The compute utilization factor is $\eta\!=\!0.9$~\cite{narayanan2021megatron}, and $T_{\mathrm{conv}}\!=\!3600$\,s/h converts the TFLOPs in $P_k^{\sf GPU}$ to an hourly capacity in~\eqref{eq:compute}. The per-token compute cost $\alpha_{i,j}^k$ is derived from the model FLOPs scaled by tier precision. The residency time $T_{i,j,k}^{\sf res}\!=\!r_i\beta_j/\text{BW}_k$ is the time a query's KV cache occupies memory. The communication delay $d_{i,j,k}^{\text{comm}}$ follows from NVLink bandwidth (600--900\,GB/s) and activation size. Per-token compute time follows the memory-bandwidth-bound decode model in~\cite{pope2023efficiently}: $d^{\sf comp}_{i,j,k}=\tau_i B_j \nu_k/\text{BW}_k$, where $\tau_i$ is a task-specific overhead, $\nu_k$ is the latency scale defined in Section~\ref{sec:model} ($\nu_k\!=\!1$ for FP16, $0.5$ for INT8, and $0.25$ for INT4), and $\text{BW}_k$ is the GPU memory bandwidth. The corresponding error multiplier $\mu_k$ in~\eqref{eq:error_calibration} is $1.0$, $1.15$, and $1.35$ for FP16, INT8, and INT4, respectively. KV-cache shapes follow the model architecture~\cite{kwon2023vllm}. This is a planning-layer analytical model: $d^{\sf comp}$ follows the standard memory-bandwidth-bound roofline for LLM decode~\cite{pope2023efficiently}, whose fidelity to measured prefill and decode latency on production GPUs has been documented in prior characterization work~\cite{patel2024splitwise,agrawal2024sarathi}. Recent high-fidelity inference simulators built on the same profiling-plus-roofline abstraction reproduce measured LLM latency to within roughly 9\% across models, GPU tiers, and parallelism configurations~\cite{agrawal2024vidur}, which is sufficient for deployment planning without bespoke per-instance measurement. The model captures the first-order feasibility drivers---per-GPU memory after sharding, the prefill/decode split under TP/PP, and quantization-induced accuracy loss. Intra-deployment runtime dynamics, including continuous-batching queueing~\cite{yu2022orca}, KV-cache contention from co-located requests, and warm-cache effects, act within a fixed deployment and are handled by the serving engine; they are therefore second-order to the placement and provisioning decision studied here. Refining the planning-layer model with a load-dependent queueing term is left to future work.

\noindent\textbf{Implementation.} All experiments use Python~3.13 and Gurobi~11~\cite{gurobi} for the exact MILP.Source code and data will be released upon acceptance.

\noindent\textbf{Comparison methods.} We benchmark the proposed GH and AGH against the exact MILP solver (DM), which yields the cost-optimal allocation wherever it terminates within the 600\,s cap, and three state-of-the-art-derived heuristic baselines introduced in Section~\ref{sec:related}: \emph{LPR}, an LP-relaxation with greedy LP-warmstart rounding; \emph{DVR}, a decoupled virtual-machine-selection-then-routing scheme after \cite{kim2025cost}; and \emph{HF}, a homogeneous-fleet provisioning scheme after DynamoLLM~\cite{stojkovic2024dynamollm}. These span the exact optimum and three dominant heuristic families: convex relaxation, decomposed VM selection, and single-tier fleet provisioning. Each baseline was originally designed for a narrower or different decision space than our joint formulation, so the comparison is not against its native problem. We adapt each one under the same network, forecast, and budget as the proposed methods to illustrate the value of enforcing coupled feasibility within a single allocation loop. The goal of this baseline study is to evaluate representative heuristic families under the common deployment space, workload assumptions, and coupled constraints considered in this paper.

\subsection{Performance Evaluation}
We employ a two-stage evaluation with $S\!=\!500$ perturbed scenarios. Each delay and error coefficient is inflated above its nominal value by up to $10$--$25\%$ (one-sided, since larger values are adverse), while each arrival rate is perturbed by $\pm20\%$.
\begin{itemize}[nosep]
    \item \textit{Stage~1 (Decision):} Each algorithm, including the exact MILP solver, computes its provisioning, placement, and parallelism decisions $(\by^*,\bz^*,\bw^*)$ using a forecasted delay/error rate. The resulting deployment decisions are then fixed.
    \item \textit{Stage~2 (Operation):} For each perturbed scenario, the Stage-1 deployment is held fixed and only the routing fractions $\bx$ and unmet-demand variables $\bu$ are re-optimized under the realized parameters. Since the deployment variables are fixed, the Stage-2 problem is a linear program and is solved exactly.
\end{itemize}
The \textit{primary metric} is the \emph{SLO violation rate}, defined as the fraction of (scenario, query-type) pairs for which more than 1\% of demand remains unserved. The \textit{secondary metric} is the \emph{expected total cost}, computed as the sum of the deterministic Stage-1 provisioning cost and the scenario-averaged Stage-2 storage, delay, and unmet-demand penalties. Under the unmet-demand cap used in the stress and rolling-horizon studies, expected cost is largely driven by the unmet-demand penalty term and therefore provides a monetary interpretation of the violation rate.

\begin{table}[t]
\caption{Stage-2 evaluation across scenarios S1--S5. Each row averages $S\!=\!500$ perturbed scenarios; bold marks the lowest cost per scenario.}\label{tab:stage2_eval}
\centering
\footnotesize
\setlength{\tabcolsep}{4pt}
\renewcommand{\arraystretch}{1.0}
\begin{tabular}{l l c c c}
\toprule
\textbf{Scenario} & \textbf{Algo.} &
\textbf{Stage-1 (\$)} &
\textbf{Cost (\$)} &
\textbf{Viol.\,(\%)} \\
\midrule
\multirow{5}{*}{\begin{tabular}[c]{@{}l@{}}
\textbf{S1:} Default \\[-1pt]
{\footnotesize $\delta\!=\!\$100,\,\phi_{\mathrm{v}}\!=\!1\!\times$}
\end{tabular}}
& GH           & 81.2           & 115.7            & \textbf{0.0}  \\
& \textbf{AGH} & 72.1           & \textbf{106.4}   & \textbf{0.0}  \\
& LPR          & 67.0           & 8749.8           & 49.3          \\
& DVR          & 66.9           & 47240.6          & 33.3          \\
& HF           & 53.6           & 4250.9           & 32.6          \\
\midrule
\multirow{5}{*}{\begin{tabular}[c]{@{}l@{}}
\textbf{S2:} Tight \\[-1pt]
{\footnotesize $\delta\!=\!\$75,\,\phi_{\mathrm{v}}\!=\!1\!\times$}
\end{tabular}}
& GH           & 63.7           & 33786.1          & \textbf{16.7} \\
& \textbf{AGH} & 53.6           & \textbf{4250.9}  & 32.6          \\
& LPR          & 67.0           & 22869.9          & 62.9          \\
& DVR          & 66.9           & 47240.6          & 33.3          \\
& HF           & 53.6           & \textbf{4250.9}  & 32.6          \\
\midrule
\multirow{5}{*}{\begin{tabular}[c]{@{}l@{}}
\textbf{S3:} Critical \\[-1pt]
{\footnotesize $\delta\!=\!\$72,\,\phi_{\mathrm{v}}\!=\!1\!\times$}
\end{tabular}}
& GH           & 63.6           & 24849.7          & 35.2          \\
& \textbf{AGH} & 53.6           & \textbf{4250.9}  & 32.6          \\
& LPR          & 67.0           & 52283.1          & 80.8          \\
& DVR          & 66.9           & 53688.8          & 49.5          \\
& HF           & 53.6           & \textbf{4250.9}  & 32.6          \\
\midrule
\multirow{5}{*}{\begin{tabular}[c]{@{}l@{}}
\textbf{S4:} Hi.\,pen. \\[-1pt]
{\footnotesize $\delta\!=\!\$75,\,\phi_{\mathrm{v}}\!=\!5\!\times$}
\end{tabular}}
& GH           & 63.7           & 168608.8         & \textbf{16.7} \\
& \textbf{AGH} & 53.6           & \textbf{5282.8}  & 32.6          \\
& LPR          & 67.0           & 23937.8          & 62.7          \\
& DVR          & 66.9           & 182043.7         & 33.3          \\
& HF           & 53.6           & \textbf{5282.8}  & 32.6          \\
\midrule
\multirow{5}{*}{\begin{tabular}[c]{@{}l@{}}
\textbf{S5:} Hi.\,pen. \\[-1pt]
+\,critical \\[-1pt]
{\footnotesize $\delta\!=\!\$72,\,\phi_{\mathrm{v}}\!=\!5\!\times$}
\end{tabular}}
& GH           & 63.6           & 111854.9         & 35.2          \\
& \textbf{AGH} & 53.6           & \textbf{5282.8}  & 32.6          \\
& LPR          & 67.0           & 113803.1         & 85.8          \\
& DVR          & 66.9           & 188517.7         & 49.6          \\
& HF           & 53.6           & \textbf{5282.8}  & 32.6          \\
\bottomrule
\multicolumn{5}{p{\columnwidth}}{\footnotesize $\phi_{\mathrm{v}}$ scales $\phi_5$ (Image Gen.) and $\phi_6$ (Video Gen.).}
\end{tabular}
\end{table}

\begin{figure}[t]
\centering
	\subfigure[Cost]{
	     \includegraphics[width=0.49\columnwidth]{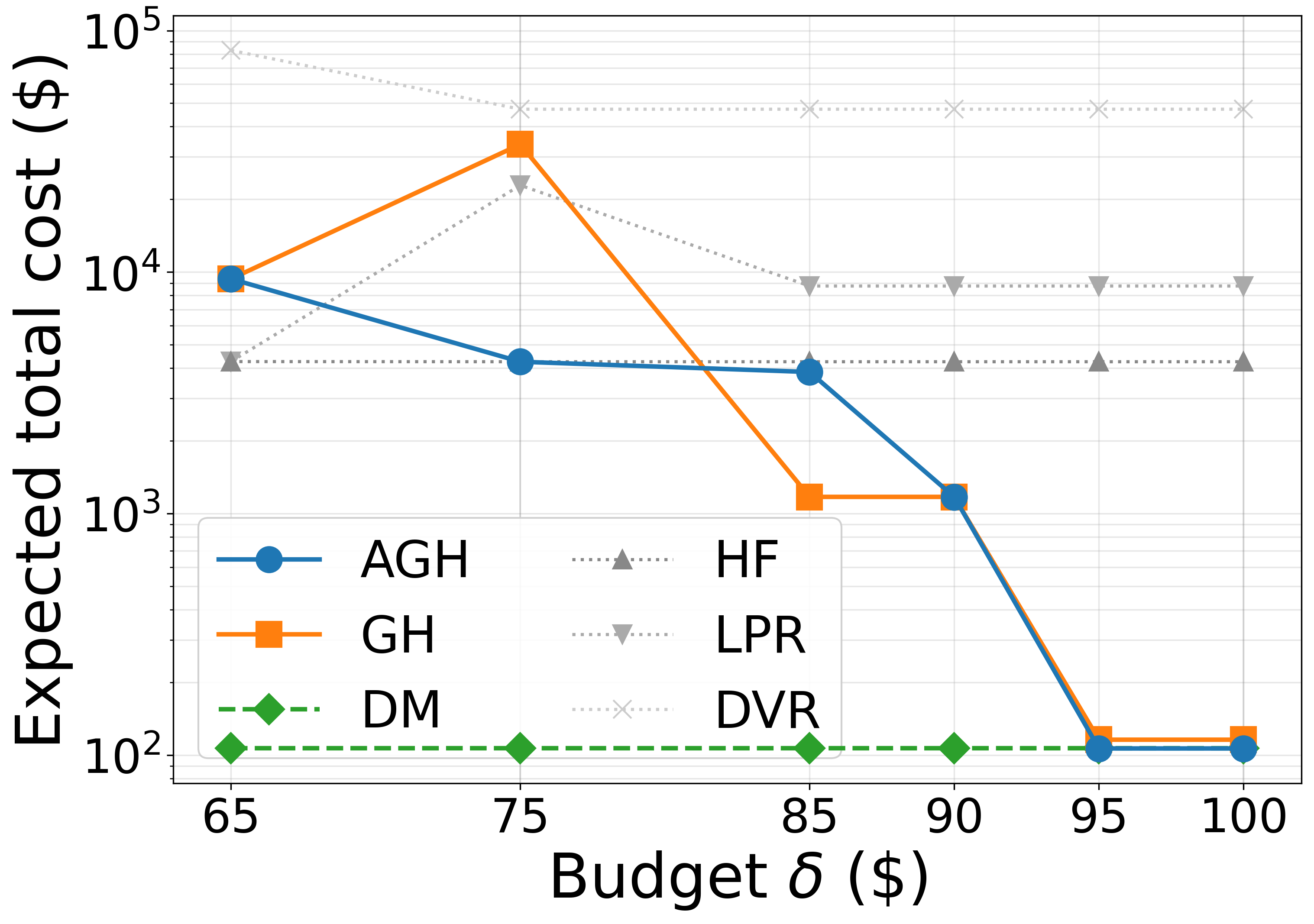}
	     \label{fig:budget_cost}
	}\hspace*{-0.5em}
	\subfigure[SLO violation]{
	     \includegraphics[width=0.49\columnwidth]{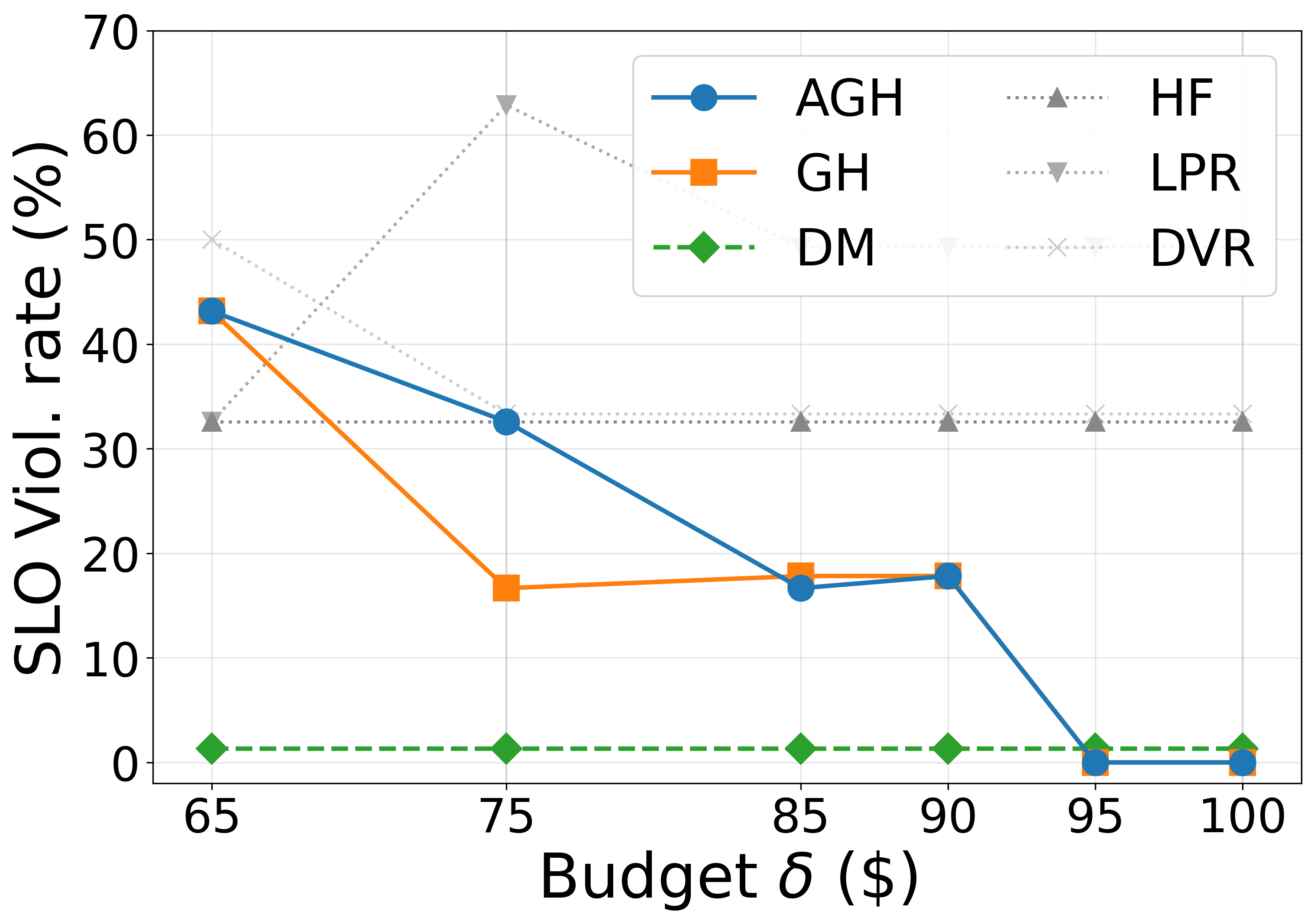}
	     \label{fig:budget_violation}
	}
    \vspace{-0.3cm}
\caption{Budget sensitivity analysis with $\phi_{\mathrm{v}}=1\times$}
\label{fig:budget_sensitivity}
\vspace{-0.5cm}
\end{figure}

\begin{figure}[t]
\centering
	\subfigure[Cost]{
	     \includegraphics[width=0.49\columnwidth]{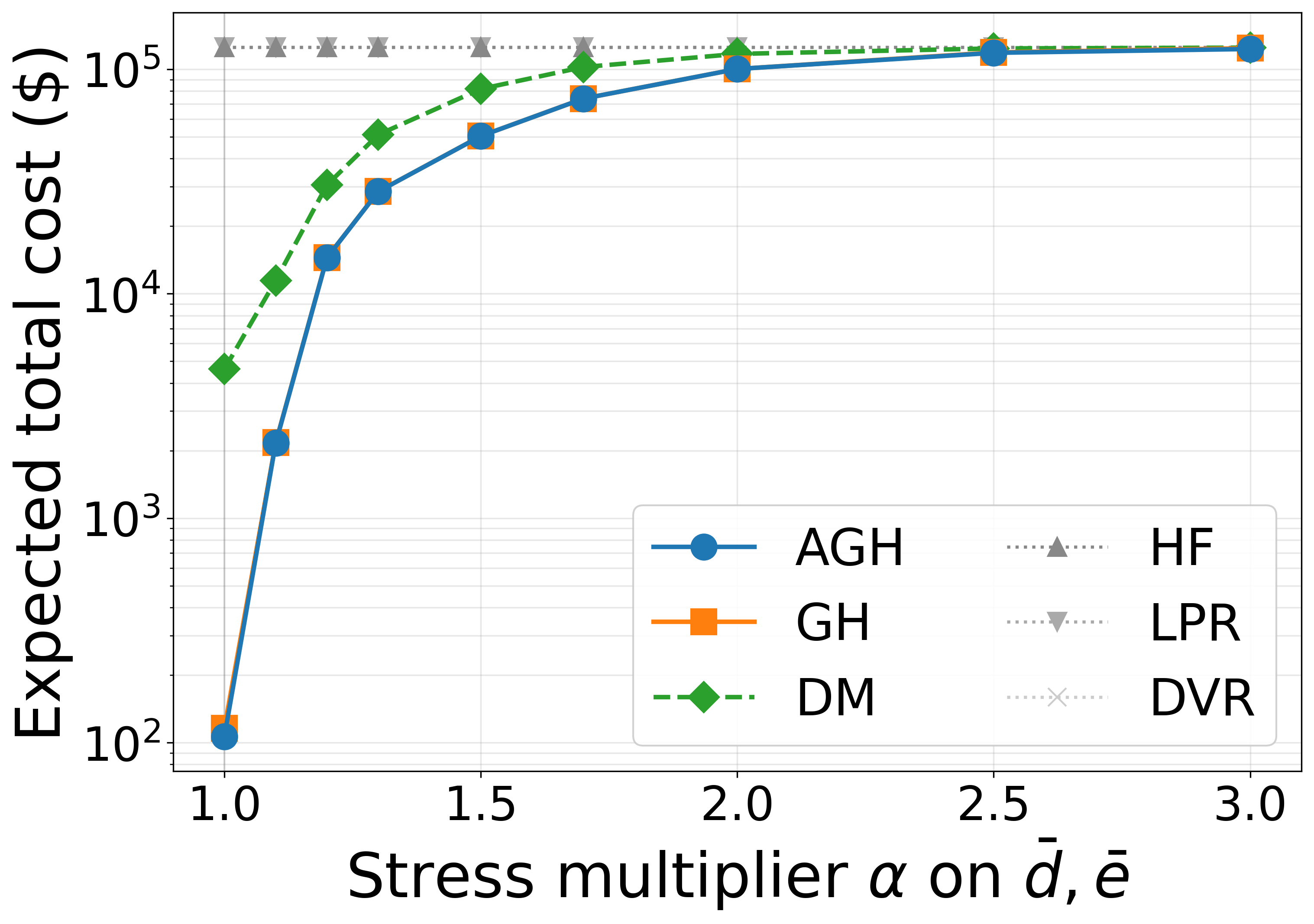}
	     \label{fig:uncertainty_cost}
	}\hspace*{-0.3em}
	\subfigure[Cost relative to DM]{
	     \includegraphics[width=0.49\columnwidth]{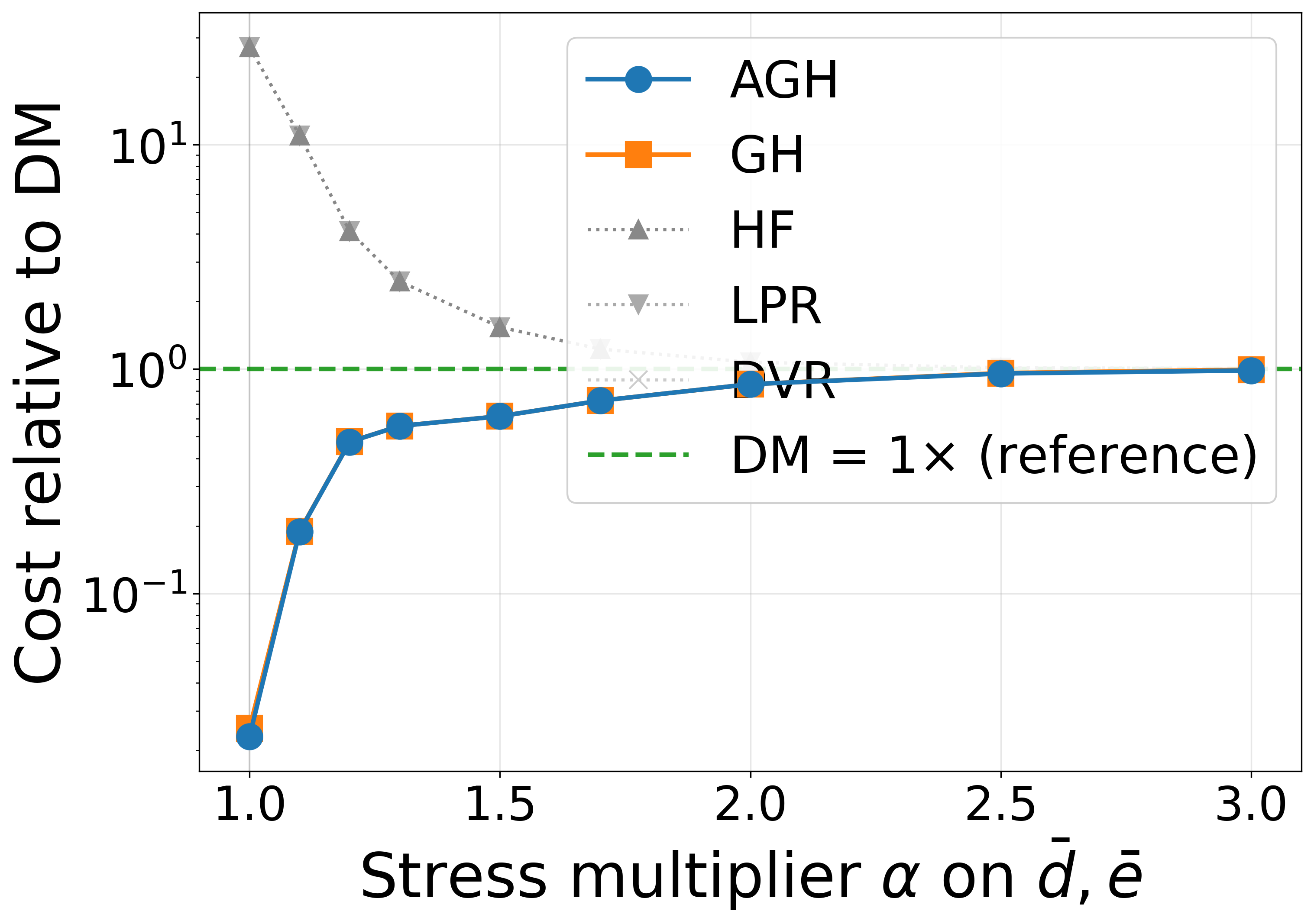}
	     \label{fig:uncertainty_ratio}
	}
    \vspace{-0.3cm}
\caption{Uncertainty-robustness study over stress multiplier $\alpha$ on $\bar d,\bar e$.}
\label{fig:uncertainty_robustness}
\vspace{-0.5cm}
\end{figure}

\begin{figure}[t]
\centering
	\subfigure[Nominal ($\alpha\!=\!1.0$)]{
	     \includegraphics[width=0.49\columnwidth]{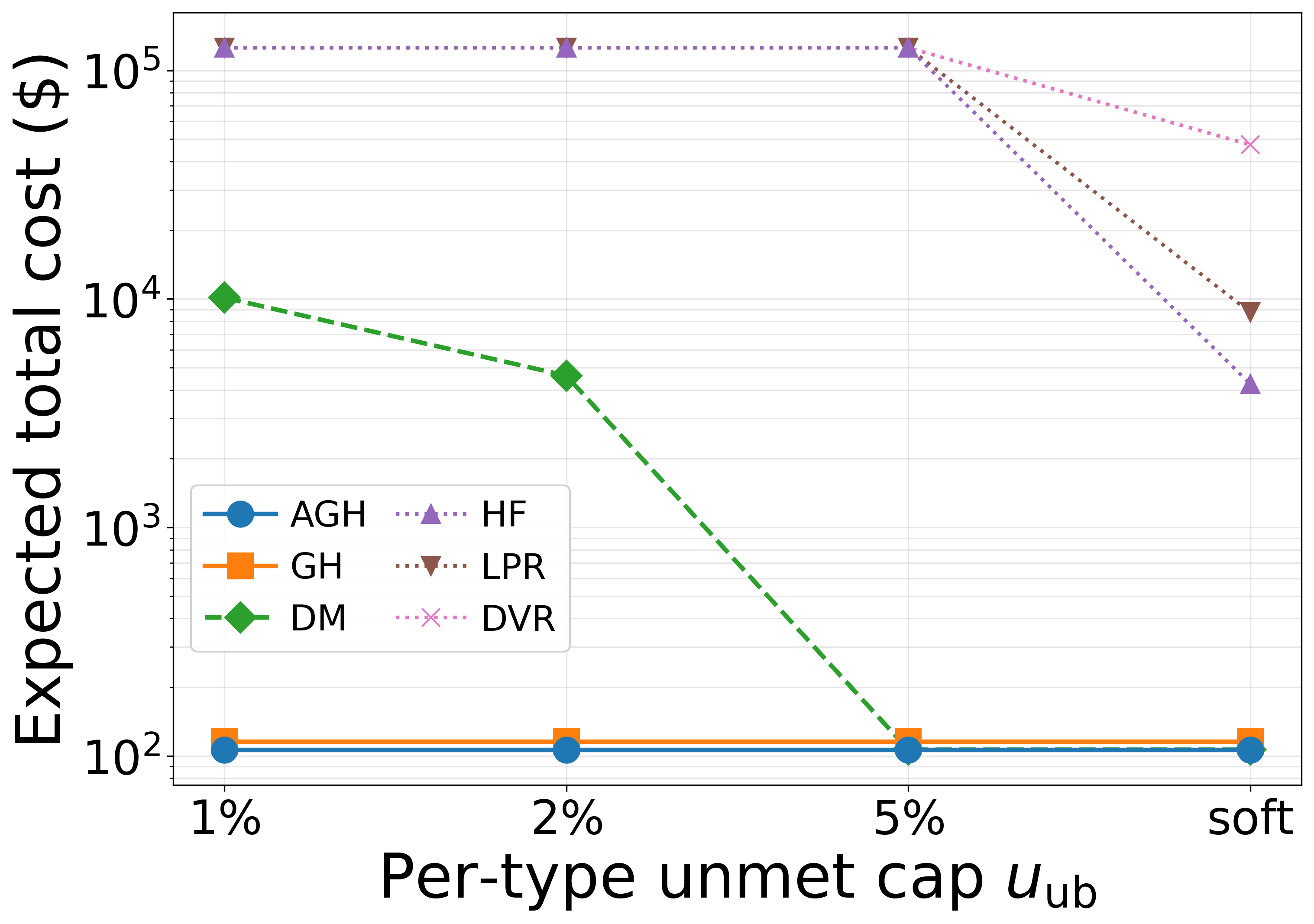}
	     \label{fig:cap_sens_nominal}
	}\hspace*{-0.5em}
	\subfigure[Stress ($\alpha\!=\!1.5$)]{
	     \includegraphics[width=0.49\columnwidth]{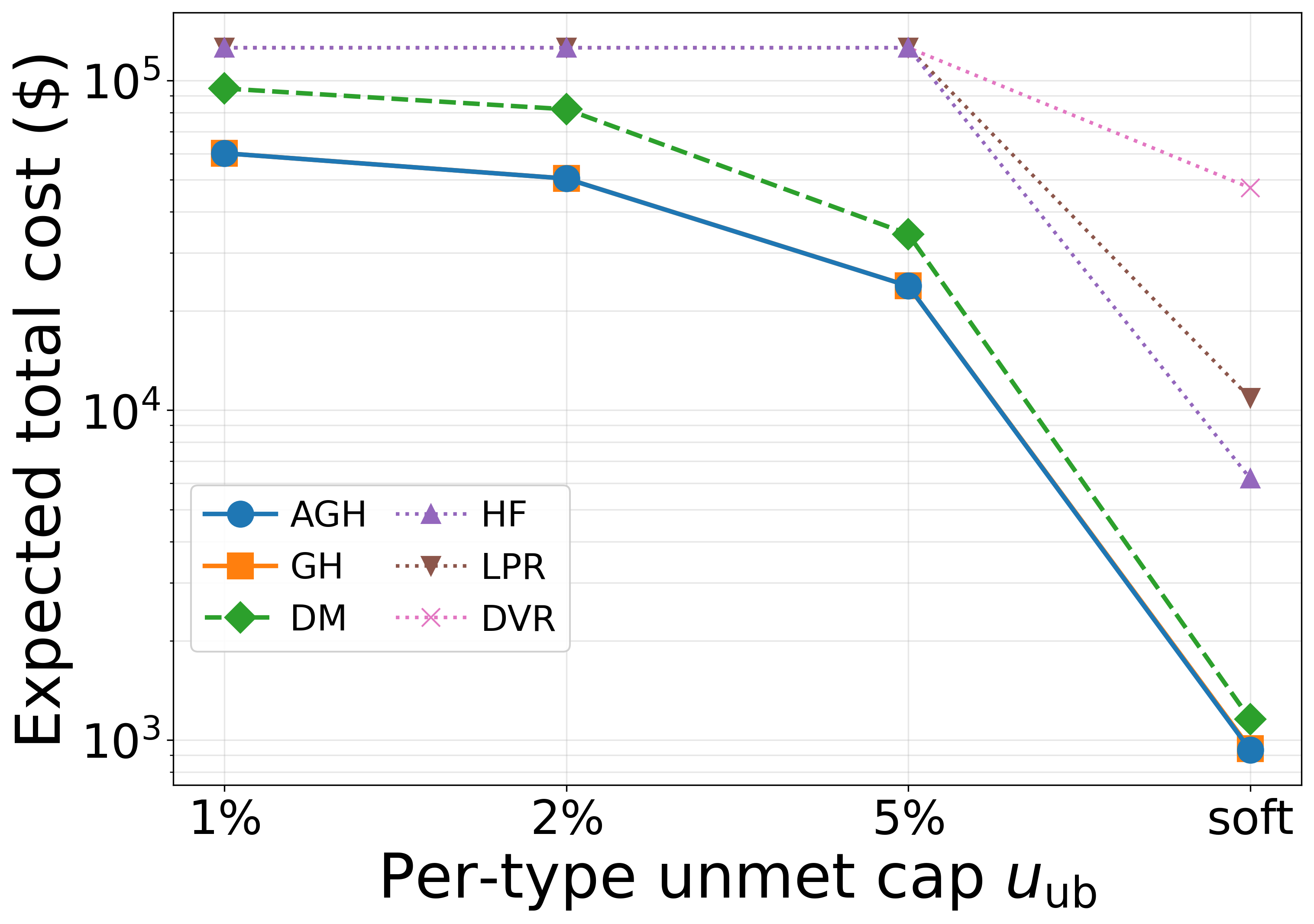}
	     \label{fig:cap_sens_stress}
	}
    \vspace{-0.3cm}
\caption{Sensitivity to the unmet cap $u_{\mathrm{ub}}$ under (a)~nominal and (b)~$1.5\times$-stressed demand. }
\label{fig:cap_sensitivity}
\vspace{-0.5cm}
\end{figure}

\noindent \textbf{Model comparison.} Table~\ref{tab:stage2_eval} compares the proposed GH and AGH against the three heuristic baselines LPR, DVR, HF across the five scenarios; Fig.~\ref{fig:budget_sensitivity} varies the budget and Fig.~\ref{fig:uncertainty_robustness} varies the uncertainty stress. Four findings stand out.

\begin{figure*}[t]
\centering
		\subfigure[Actual cost: stress test]{
	     \includegraphics[width=0.30\textwidth,height=0.15\textheight]{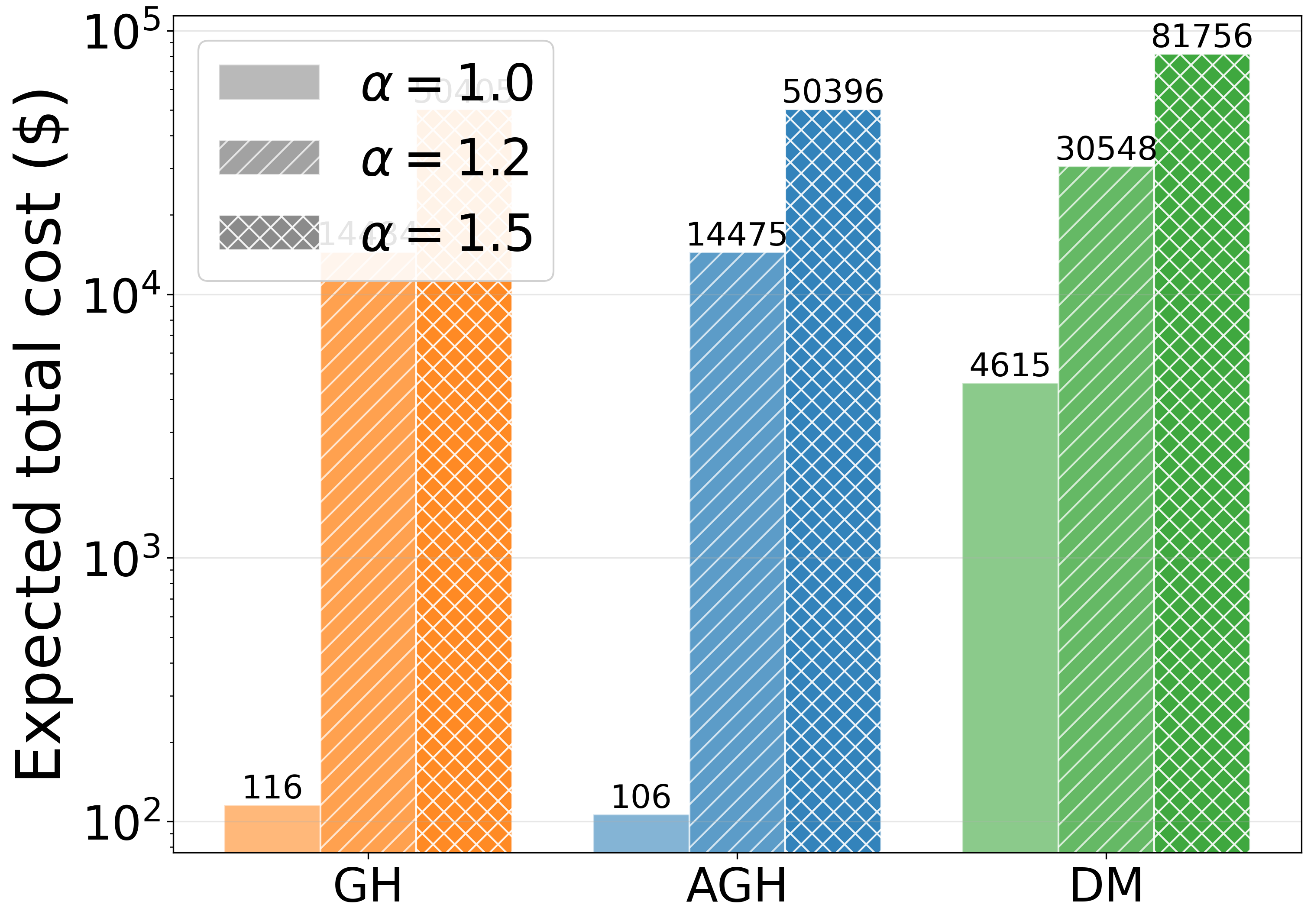}
	     \label{fig:stress_cost}
	}  
	     \subfigure[SLO violation: stress test]{
	     \includegraphics[width=0.3\textwidth,height=0.15\textheight]{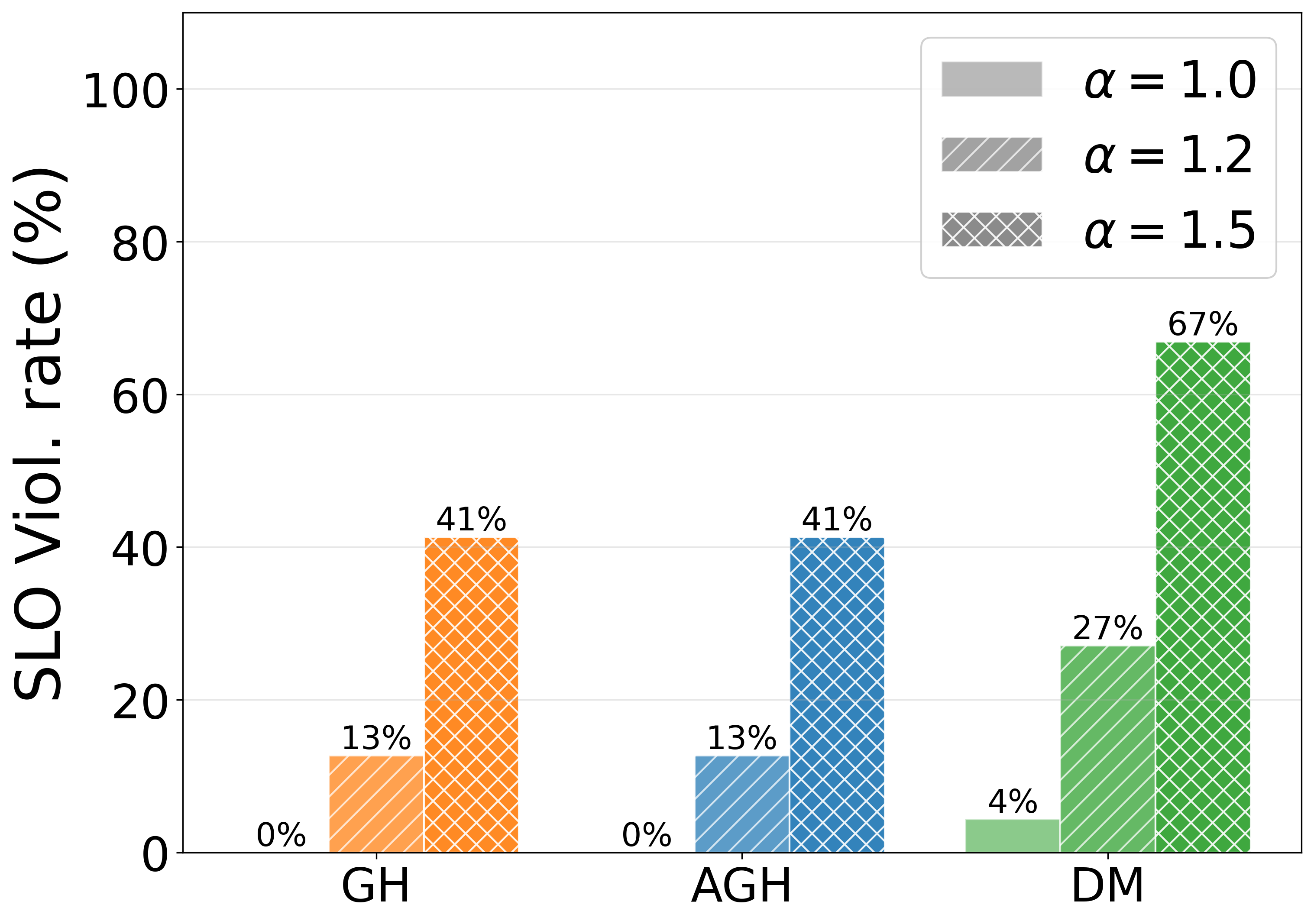}
	     \label{fig:stress_violation}
	}
	\subfigure[Cost breakdown: stress test]{
	     \includegraphics[width=0.3\textwidth,height=0.15\textheight]{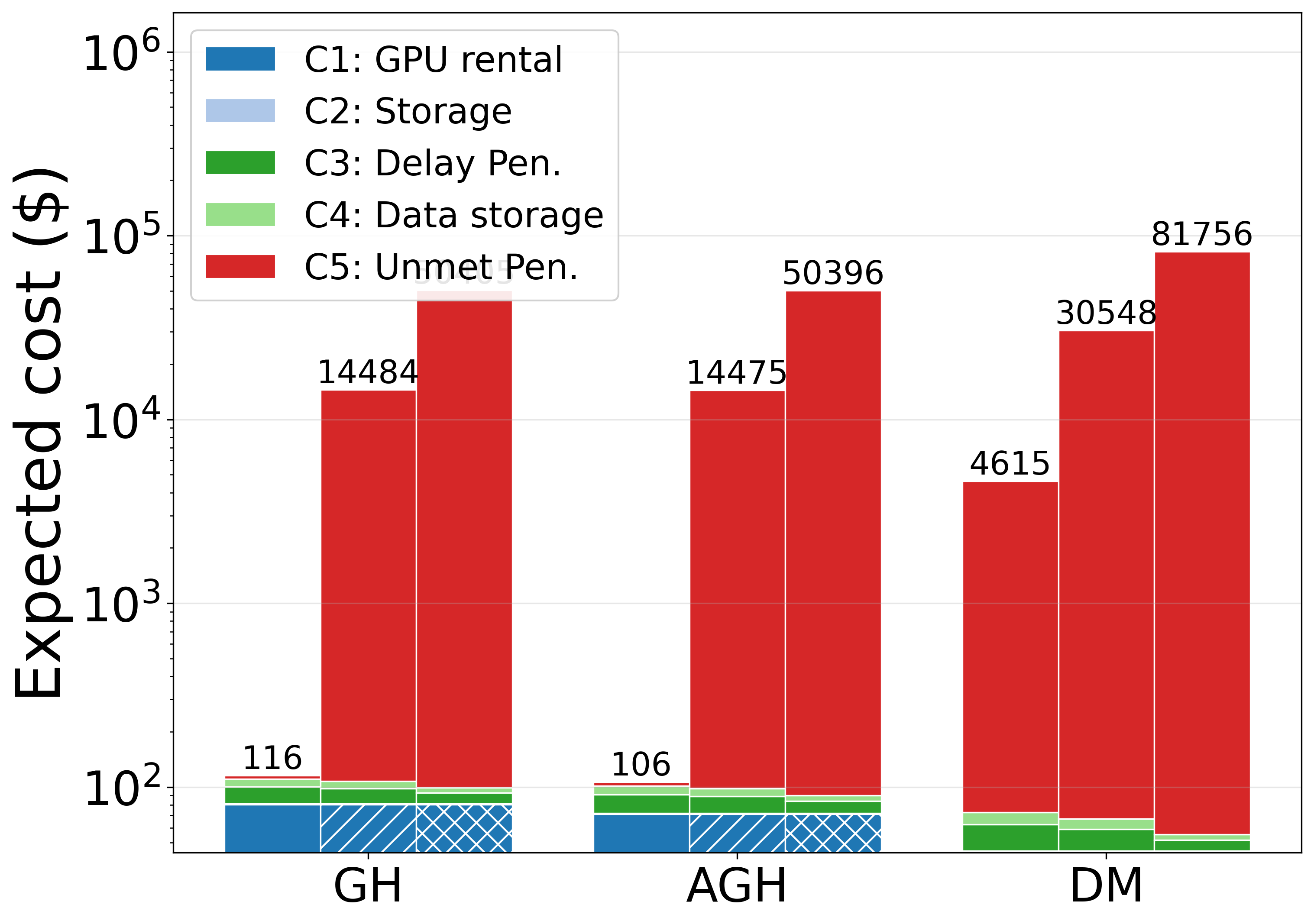}
	     \label{fig:cost_breakdown}
	}
	\subfigure[Delay SLO $\Delta_i$ and accuracy tolerance $\epsilon_i$]{
	     \includegraphics[width=0.3\textwidth,height=0.15\textheight]{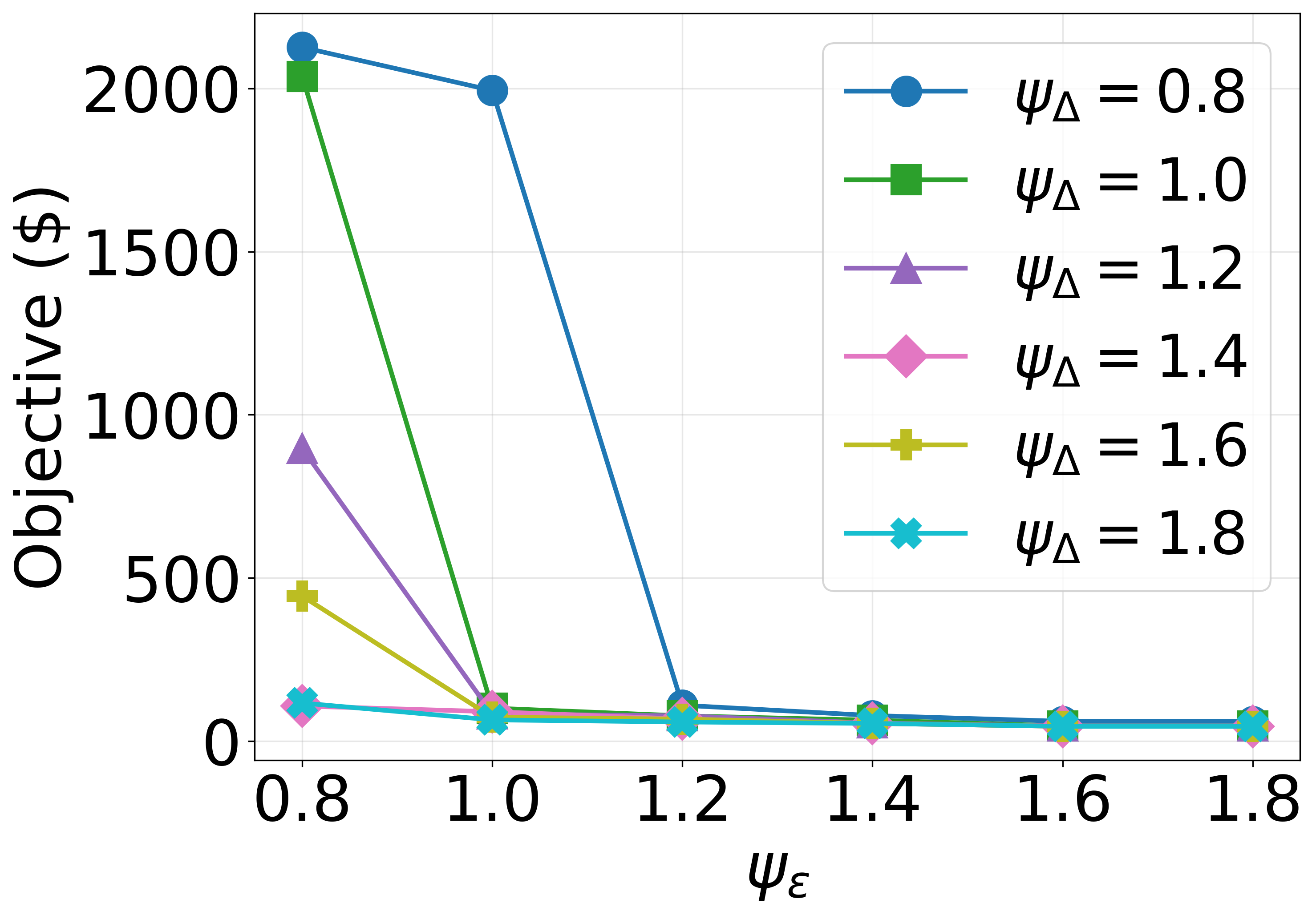}
	     \label{fig:vary_delay_error}
	}  
	     \subfigure[Budget $\delta$ and GPU rental price $p_k^c$]{
	     \includegraphics[width=0.3\textwidth,height=0.15\textheight]{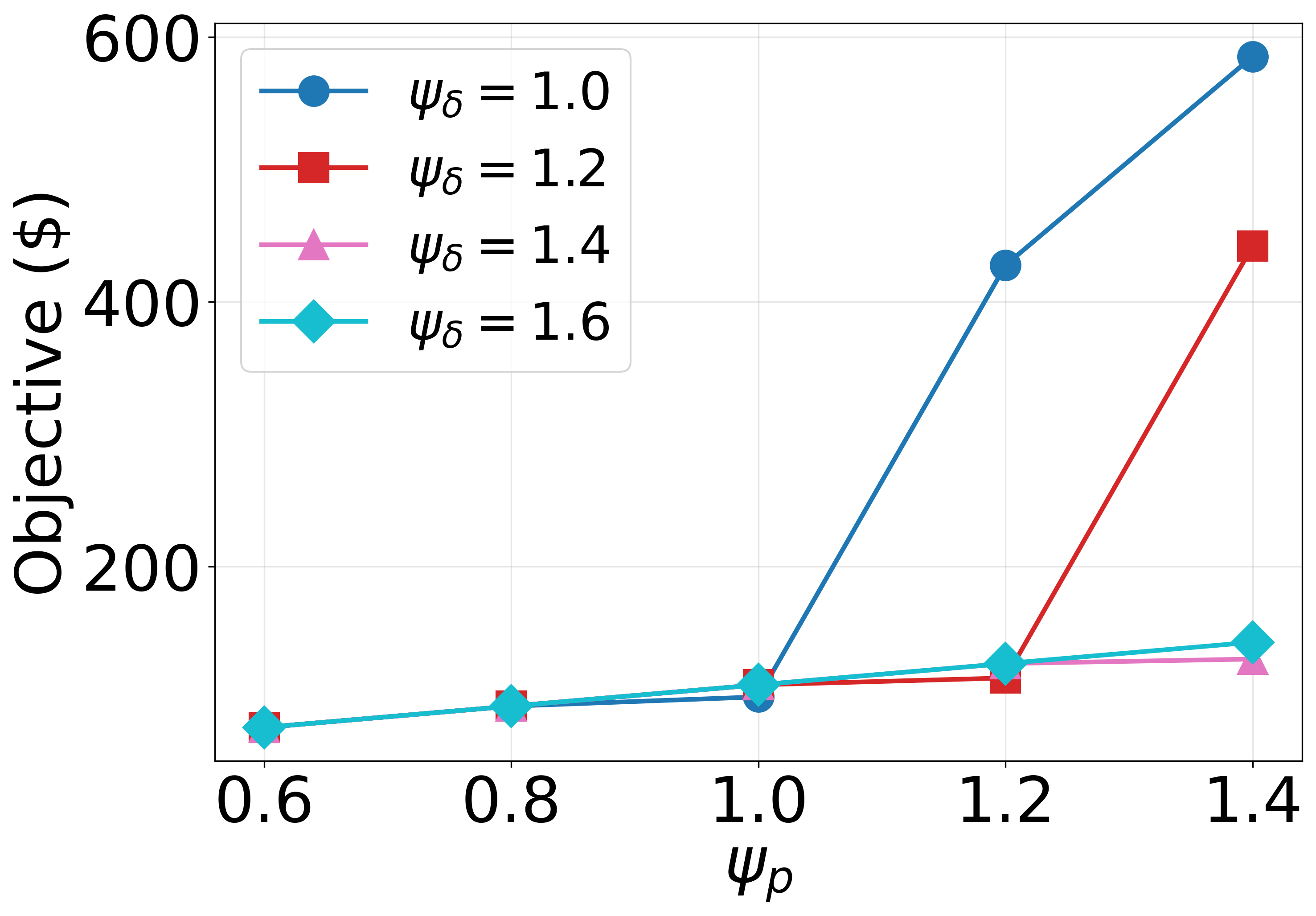}
	     \label{fig:vary_budget_rental}
	}
	\subfigure[GPU rental price $p_k^c$ and delay SLO $\Delta_i$]{
	     \includegraphics[width=0.3\textwidth,height=0.16\textheight]{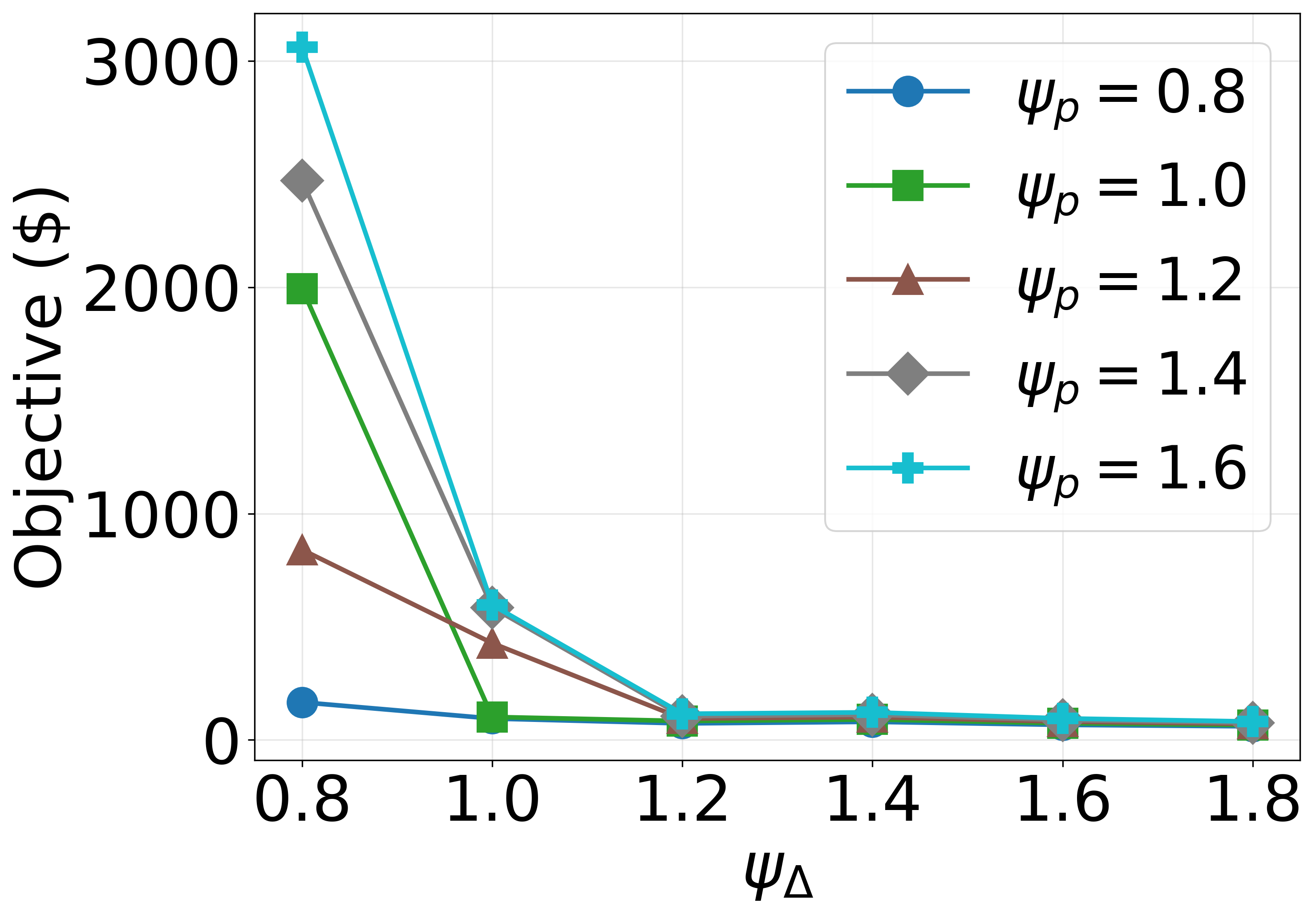}
	     \label{fig:vary_gpu_delay}
	}
	\caption{Stress and sensitivity study (strict cap $u_i\!\le\!0.02$). (a)--(c) GH, AGH, DM under $1.0/1.2/1.5\times$ delay--error inflation: the heuristics' headroom keeps demand served, while the cost-minimal exact plan breaches the cap, so its unmet penalty ($\mathcal{C}_5$) dominates cost (panel (c)). (d)--(f) AGH variations over $\Delta_i$, $\epsilon_i$, $\delta$, $p_k^c$.}
\end{figure*}

(a) Under the \emph{default setting (S1)}, AGH achieves the lowest expected cost (\$106.4) and zero SLO violations, outperforming all external baselines, whose violation rates range from 32.6\% to 49.3\%. GH achieves the same zero-violation rate with a slightly higher cost (\$115.7).

(b) \emph{Under tighter budgets (S2--S5)}, AGH consistently outperforms LPR and DVR in both cost and violation rate. In these scenarios, AGH converges to the same H100-INT8 single-pair deployment as HF and therefore achieves identical costs. Compared with LPR and DVR, AGH reduces expected cost by factors of 5--11$\times$ while maintaining substantially lower violation rates.

(c)~\emph{GH exhibits a different cost–reliability tradeoff.} Under the tight-budget scenarios S2 and S4, GH achieves the lowest violation rate among the heuristic methods, but at a substantially higher expected cost than AGH. Under the more restrictive scenarios S3 and S5, GH's violation rate increases to 35.2\%, and its expected cost rises sharply because the unmet-demand penalty dominates the objective. By contrast, LPR and DVR violate at $\geq\!33\%$ in every tight scenario \emph{and} cost more than AGH since the decomposed and convex-relaxation baselines do not enforce the coupled feasibility our design targets. AGH optimizes expected cost rather than violation rate alone, and therefore accepts modest increases in violations when doing so substantially reduces unmet-demand penalties and provisioning cost. This explains why AGH may exhibit slightly higher violation rates than GH in some budget-constrained scenarios while achieving significantly lower overall cost.

(d)~\emph{Robustness under uncertainty}). Fig.~\ref{fig:uncertainty_robustness} evaluates robustness under increasing delay and error inflation. As uncertainty grows beyond the planning forecast, AGH and GH maintain substantially lower costs than the external baselines, while HF, LPR, and DVR rapidly approach the fully-unserved regime. The advantage of AGH and GH is most pronounced under moderate uncertainty and narrows only at the highest stress levels, where the available provisioning slack of any fixed deployment becomes insufficient.

These results suggest that optimality alone does not guarantee robust operational performance. Although the exact MILP minimizes cost for the forecasted parameters, its solutions may become sensitive to their deviations. In contrast, the decisions from GH and AGH maintain larger feasibility margins, resulting in more gradual degradation of cost and SLO performance under uncertainty.

\noindent \textbf{Ablation.} Table~\ref{tab:ablation} disables each mechanism in isolation on the default setup. Disabling M1 (parallelism-aware feasibility filter) lets the cost-only ranker greedily select cheap tiers that cannot hold the chosen model under the candidate TP/PP or whose TTFT alone already exceeds $\Delta_i$; the budget is exhausted on inadmissible placements and the construction terminates infeasible (memory violation). Disabling M3 (parallelism upgrade on active pairs) forces every TP increase to activate a fresh pair, paying the full activation cost and duplicating model weights, so the budget is consumed faster and late-arriving queries land on residual capacity at a TP too low to meet $\Delta_i$ (delay violation). Disabling M2 (cost-per-effective-coverage ranking) is different in kind: with M1 and M3 still active the plan remains feasibility-safe, but raw-cost ranking commits to cheap-but-narrow candidates (e.g., a low-cost INT4 tier whose error budget admits only a small fraction of a strict-accuracy query type) and forces additional pair activations to cover the residual, inflating cost by nearly 50\%. M1 and M3 are therefore feasibility prerequisites; M2 is a quality optimization.

\begin{table}[t]
\caption{Ablation of the three constraint-aware mechanisms.}
\label{tab:ablation}
\centering
\footnotesize
\setlength{\tabcolsep}{4pt}
\begin{tabular}{lcc}
\toprule
\textbf{Configuration} & \textbf{Feasible?} & \textbf{Cost (\$)} \\
\midrule
AGH (all M1--M3)              & Yes & 89.88 \\
w/o M1 (feasibility filter)   & No (memory)  & n/a \\
w/o M2 (cost-only ranking)    & Yes & 134.52 ($+50\%$) \\
w/o M3 (TP upgrade)           & No (delay)   & n/a \\
\bottomrule
\end{tabular}
\end{table}

\noindent \textbf{Sensitivity analysis.} 
Figs.~\ref{fig:cap_sensitivity}--\ref{fig:vary_gpu_delay} evaluate the sensitivity of the proposed framework to key system parameters and design assumptions.

\noindent\textbf{Unmet-demand cap.}
The stress study uses a strict per-type unmet-demand cap of $u_i\!\le\!2\%$, corresponding to a requirement that at least $98\%$ of each query type's demand be served. Fig.~\ref{fig:cap_sensitivity} repeats the S1 evaluation for $u_{\mathrm{ub}}\!\in\!\{1\%,2\%,5\%,\text{soft}\}$. The relative ordering of the methods remains unchanged across all cap values. Relaxing the cap primarily reduces the unmet-demand penalty and therefore shifts the absolute cost level without materially affecting comparative performance. Under normal demand and a soft cap, the exact MILP recovers the lowest-cost solution, while AGH remains within $0.3\%$ of the optimum. Under stressed conditions, however, GH and AGH consistently outperform the exact MILP, indicating that their advantage stems from maintaining additional feasibility margin under uncertainty rather than from the cap setting itself.

\noindent \textbf{\emph{Delay SLO versus accuracy tolerance}} (Fig.~\ref{fig:vary_delay_error}). Tightening the delay SLO $\Delta_i$ forces higher TP degrees: in the bandwidth-bound decode regime, per-token compute time scales as $d^{\sf comp}/\text{TP}$, so the per-pair GPU count grows roughly with $1/\Delta_i$ until either the per-tier TP cap is hit or the candidate is discarded by mechanism M1. Tightening the error tolerance $\epsilon_i$ instead acts as a discrete admissibility filter: the precision-keyed multiplier $\mu_k$ takes only three values, so below a query type's FP16 admission threshold all INT8 and INT4 placements are removed from the candidate set in one step. The two effects compose, but the delay axis dominates across the tested range because tightening $\Delta_i$ walks the parallelism ladder one rung at a time, while $\epsilon_i$ only flips discrete tiers in or out.

\noindent \textbf{\emph{Budget versus rental price}} (Fig.~\ref{fig:vary_budget_rental}). Rising $p_k^c$ changes AGH's topology, not just its absolute scale. As prices rise, the cost of activating a fresh (model, tier) pair grows faster than the cost of adding GPUs to an already-loaded pair, so the planner consolidates onto fewer, higher-capacity tiers (H100/A100 in place of A6000/RTX 4090) and packs more query types per pair via mechanism M3. Tightening the budget $\delta$ amplifies this consolidation; AGH reaches the budget frontier by enlarging the coverage set per active pair rather than by downgrading tiers.

\noindent \textbf{\emph{Rental price versus delay SLO}} (Fig.~\ref{fig:vary_gpu_delay}). The two parameters interact rather than simply add. Relaxing $\Delta_i$ allows lower TP, which lowers the per-pair GPU count and dampens sensitivity to $p_k^c$ in absolute terms: in the relaxed-delay regime, AGH's cost becomes near-flat in the rental price because the planner has already converged on the cheapest feasible TP. Under a tight delay SLO, the same rental-price increase translates almost linearly into total cost. The operational implication is that, when $p_k^c$ is high, relaxing the delay SLO is a more effective cost-reduction lever than downgrading the tier or the precision.

Overall, the results show that delay requirements primarily drive parallelism and GPU demand, whereas accuracy requirements act as feasibility filters that determine which model--tier configurations remain admissible.

\noindent\textbf{Runtime scaling.}
Table~\ref{tab:scalability} reports runtime as the problem size increases. The exact MILP remains below 14\,s up to $(I,J,K)\!=\!(10,10,10)$, but exceeds the 600\,s time limit at $(15,15,10)$ and does not terminate within the limit at $(20,20,20)$. In contrast, GH remains below 1\,s on all tested instances ($\leq0.9$\,s at $(20,20,20)$), while AGH remains below 3\,s ($\leq2.3$\,s at $(20,20,20)$). Relative to the MILP time limit, AGH therefore achieves a speedup of at least $260\times$ on the largest instance. This value is a lower bound because the MILP does not terminate within the prescribed time limit. On instances where the MILP terminates, AGH remains within a few percent of the optimal objective value. For larger instances, where the optimum is unavailable, we report only runtime, feasibility, and the robustness results presented earlier. These results indicate that AGH retains solution quality on moderate-sized instances while scaling to problem sizes that are impractical for exact optimization.

\noindent\textbf{Operational target.} 
The target control interval in our study is the 5-minute rolling re-optimization horizon considered in Section~\ref{sec:rolling}. At this timescale, both GH and AGH complete well within the available decision window, enabling deployment updates without disrupting inference operation. GH provides sub-second response times across all tested instances, while AGH offers improved solution quality with runtime below 3\,s, making both methods suitable for online deployment in dynamic LLM-serving environments.

\subsection{Rolling-Horizon Adaptation}
\label{sec:rolling}

\begin{figure}[t]
\centering
	\subfigure[Static]{
	     \includegraphics[width=0.48\columnwidth]{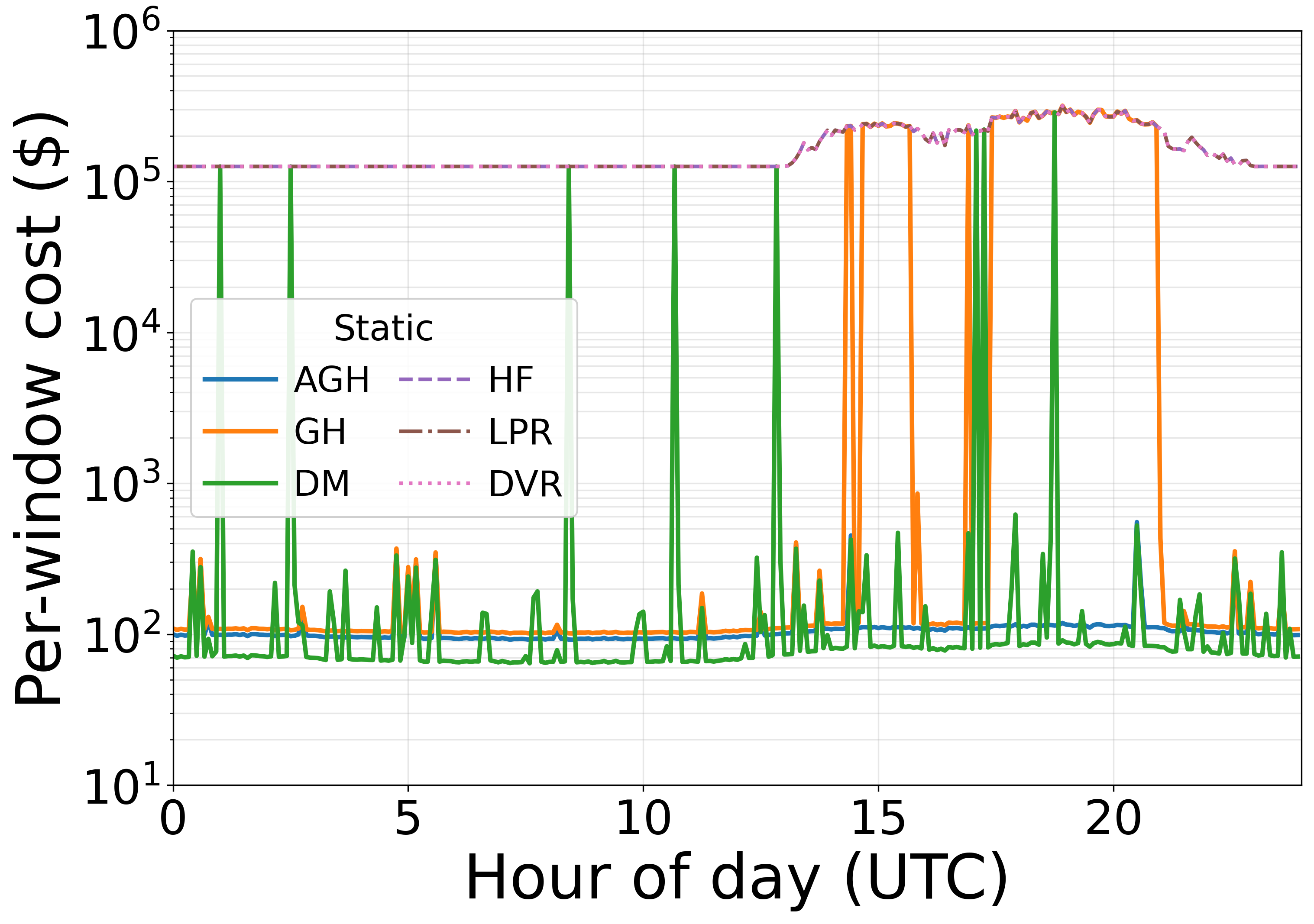}
	     \label{fig:trace_cost_static}
	}\hspace*{-0.5em}
	\subfigure[5-min rolling]{
	     \includegraphics[width=0.48\columnwidth]{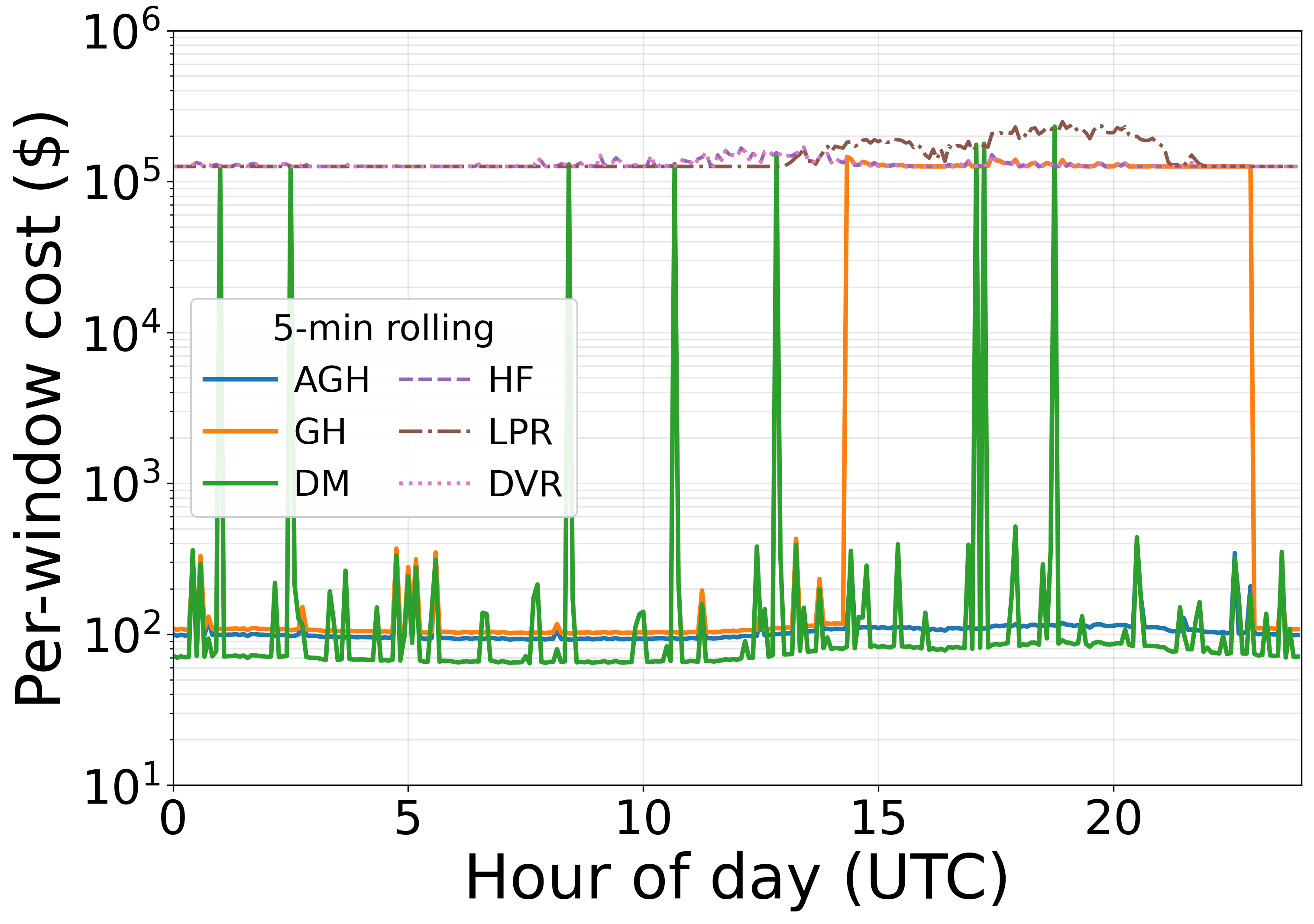}
	     \label{fig:trace_cost_5min}
	}
\caption{Per-window cost on the Azure code trace, shown separately for (a)~static method and (b)~5-minute rolling horizon.}
\label{fig:trace_cost}
\end{figure}

The few-second runtime of GH and AGH makes rolling re-optimization feasible at operational timescales. We evaluate this capability in two complementary settings: a synthetic demand-volatility study based on a geometric random walk (Table~\ref{tab:rolling}) and a replay of a real Azure trace (Table~\ref{tab:trace}, Fig.~\ref{fig:trace_cost}).

\noindent\textbf{Synthetic volatility study.}
We divide the 24-hour horizon ($\Delta_T\!=\!24$\,h) into 288 five-minute windows and evolve demand as $\lambda_i^{(t+1)}=\lambda_i^{(t)}\exp(\mathcal{N}(0,\sigma))$, where $\sigma$ is the per-step volatility. We test $\sigma\in\{0.01,0.02,0.03,0.04,0.05\}$, corresponding to cumulative demand standard deviations of approximately $17\%$, $34\%$, $51\%$, $68\%$, and $85\%$ over the horizon. Static methods (\mbox{DM-24h}, \mbox{GH-24h}, \mbox{AGH-24h}) solve once at $t=0$, whereas \mbox{AGH-5min} re-solves every five minutes using a keep-best rule. Table~\ref{tab:rolling} reports the mean cost over 30 independent trials.

Table~\ref{tab:rolling} shows that AGH-24h achieves the lowest cost across all volatility levels. For $\sigma\le0.04$, AGH-5min and AGH-24h produce nearly identical results, indicating that the static allocation already absorbs most demand fluctuations. At $\sigma=0.05$, both methods experience degradation as cumulative demand drift exceeds the available provisioning margin. GH remains competitive only at low volatility and deteriorates more rapidly as demand drift increases, whereas the exact MILP remains substantially more expensive throughout the tested range.

\noindent\textbf{Real Azure trace.}
The geometric random walk provides a memoryless stress test; we therefore complement it with a replay of the Azure LLM code-completion trace~\cite{azurellmtrace2025} to evaluate rolling re-optimization under structured demand variation. The trace contains approximately 16.8 million requests collected between 2024-05-10 and 2024-05-16. We use the busiest day, 2024-05-14 (3.21 million requests), and aggregate requests into 288 five-minute windows. The resulting request rate varies by approximately $10\times$ between the early-morning trough ($\sim28$\,k/h) and the evening peak ($\sim300$\,k/h). We treat the per-window request count as a diurnal multiplier relative to the daily average and scale each query type's nominal arrival rate accordingly.

To evaluate the effect of rolling re-optimization across all methods, we consider both a \emph{static} variant, which solves Stage~1 once using the day-average forecast, and a \emph{5min} variant, which re-optimizes every five minutes using an EWMA forecast and adopts the new configuration only if it improves upon the incumbent. The same re-optimization protocol is applied to all methods, ensuring that any difference between the static and rolling variants is attributable to the deployment produced by the method rather than to the re-optimization procedure itself.


\begin{table}[t]
\centering
\caption{Rolling-horizon cost under a \emph{synthetic} geometric-random-walk volatility study (\$, mean over the 24~h horizon, 30 trials). AGH-24h is the cheapest method at every $\sigma$. }
\label{tab:rolling}
\resizebox{\columnwidth}{!}{%
\begin{tabular}{l ccccc}
\toprule
 & $\sigma\!=\!0.01$ & $\sigma\!=\!0.02$ & $\sigma\!=\!0.03$ & $\sigma\!=\!0.04$ & $\sigma\!=\!0.05$ \\
\midrule
DM-24h    & $4139$ & $4144$ & $4152$ & $4164$ & $4179$ \\
GH-24h    & $118$  & $118$  & $165$  & $3192$ & $9589$ \\
GH-5min   & $118$  & $118$  & $284$  & $6277$ & $10768$ \\
AGH-24h   & $108$  & $109$  & $109$  & $109$  & $1036$ \\
AGH-5min  & $108$  & $109$  & $109$  & $109$  & $1723$ \\
\midrule
\multicolumn{1}{r}{\footnotesize \emph{AGH-5min vs AGH-24h}} &
  \footnotesize $0.0\%$ &
  \footnotesize $0.0\%$ &
  \footnotesize $0.0\%$ &
  \footnotesize $0.0\%$ &
  \footnotesize $\!+\!66.3\%$ \\
\multicolumn{1}{r}{\footnotesize \emph{AGH-24h vs DM-24h}} &
  \footnotesize $-97.4\%$ &
  \footnotesize $-97.4\%$ &
  \footnotesize $-97.4\%$ &
  \footnotesize $-97.4\%$ &
  \footnotesize $-75.2\%$ \\
\bottomrule
\end{tabular}}

\parbox{\columnwidth}{\footnotesize \emph{Protocol:} each trial spans 288 5-minute windows; demand follows a geometric random walk $\lambda^{(t+1)}\!=\!\lambda^{(t)}\exp(\mathcal{N}(0,\sigma))$; the per-window Stage-2 LP uses the strict $u_i\!\leq\!0.02$ per-type unmet cap, matching the stress protocol of Fig.~\ref{fig:stress_cost}. The $\sigma\!=\!0.05$ means are tail-driven---a minority of trials whose cumulative drift exhausts any fixed plan's headroom---so the medians remain near \$108--118 for AGH/GH.}
\end{table}

\begin{table}[t]
\caption{Rolling-horizon cost on the \emph{real} Azure diurnal trace}
\label{tab:trace}
\centering
\footnotesize
\setlength{\tabcolsep}{5pt}
\begin{tabular}{l r r r}
\toprule
\textbf{Method} & \textbf{Mean cost/win (\$)} & \textbf{Total 24 h (\$)} & \textbf{Viol.\,(\%)} \\
\midrule
\textbf{AGH-5min} & \textbf{111.7}  & \textbf{32{,}159.7}     & \textbf{0.0}  \\
AGH-static        & 112.9           & 32{,}503.0              & 0.0           \\
\midrule
DM-5min           & 4{,}440.6       & 1{,}278{,}879.1         & 6.6           \\
DM-static         & 4{,}800.6       & 1{,}382{,}565.0         & 6.6           \\
\midrule
GH-5min           & 46{,}357.1      & 13{,}350{,}852.8        & 36.1          \\
GH-static         & 53{,}798.2      & 15{,}493{,}877.0        & 21.2          \\
\midrule
HF-5min           & 130{,}380.1     & 37{,}549{,}463.2        & 100.0         \\
HF-static         & 165{,}323.0     & 47{,}613{,}027.1        & 100.0         \\
LPR-5min          & 146{,}664.3     & 42{,}239{,}306.3        & 100.0         \\
LPR-static        & 165{,}336.4     & 47{,}616{,}885.3        & 100.0         \\
DVR-5min          & 130{,}430.5     & 37{,}563{,}987.8        & 100.0         \\
DVR-static        & 165{,}336.3     & 47{,}616{,}864.2        & 100.0         \\
\bottomrule
\end{tabular}
\end{table}

Three observations emerge. First, AGH's \emph{static} plan is the most robust: AGH-24h holds near \$108--109 with negligible variance for $\sigma\!\le\!0.04$ and remains the cheapest method at every volatility level. Only at $\sigma\!=\!0.05$ does its mean rise (to \$1{,}036), and that rise is tail-driven---a minority of trials whose cumulative drift eventually exhausts any fixed plan's headroom (the median stays \$108). Second, rolling re-optimization yields \emph{no} mean-cost reduction in this random-walk setting: AGH-5min reproduces AGH-24h exactly for $\sigma\!\le\!0.04$ and never improves on it, because AGH's headroom-aware Stage-1 plan already absorbs the drift, leaving nothing for re-solving to recover; the benefit of rolling instead surfaces as peak-window smoothing on the real diurnal trace (Table~\ref{tab:trace}, Fig.~\ref{fig:trace_cost}), not as mean reduction under a memoryless walk. Third, the exact MILP is consistently dominated: DM-24h costs \$4{,}100--4{,}200 across all $\sigma$ because its cost-minimal, headroom-free placement pays the per-type unmet penalty whenever realized parameters exceed the forecast. GH is robust only up to $\sigma\!=\!0.02$ and degrades sharply thereafter (\$3{,}192 at $\sigma\!=\!0.04$), confirming that AGH's multi-start construction provisions markedly more drift-absorbing headroom than GH's single-pass ordering.

The exact MILP optimizes against the forecasted demand profile and does not explicitly account for out-of-sample demand variation. Consequently, under large demand surges the unmet-demand penalty dominates the realized operational cost, leading to substantially higher cost than AGH despite its optimality under the forecast. Taken together, the synthetic and trace-driven studies indicate that the primary performance gain comes from the quality of the deployment itself rather than from frequent re-optimization. Rolling re-optimization provides additional adaptability and reduces peak-window cost, but its benefit is limited when the static allocation already contains sufficient capacity to absorb demand variation.

\noindent\textbf{Generalization across days.} To assess whether these observations are specific to a single day, we repeat the experiment on a second, more volatile day from the same trace (2024-05-15), whose peak-to-trough demand ratio is $15.6\times$ compared with $10\times$ on 2024-05-14. The results are highly consistent across the two days: AGH maintains a 24-hour cost of approximately \$32.4\,k with $0\%$ violations, the exact MILP incurs approximately \$1.40\,M with $6.6\%$ violations, GH reaches \$16.4\,M, and the external baselines remain near \$47.7\,M with $100\%$ violations. Across all methods, the key performance metrics vary by less than $2\%$ between days (less than $6\%$ for GH), indicating that the relative method ranking and the principal conclusions remain stable despite the higher demand volatility.

\begin{table}[t]
\caption{Runtime scaling with problem size $(I,J,K)$, in seconds. DM exceeds the 600\,s cap at $(15,15,10)$; GH stays under 1\,s and AGH under 3\,s on all tested instances (speedup lower bound $\geq\!260\times$ at $(20,20,20)$).}
\centering
\label{tab:scalability}
\setlength{\tabcolsep}{4pt} 
\footnotesize 
\begin{tabular}{lccccc}
\toprule
\textbf{Method}
& $(4,4,5)$
& $(6,6,10)$
& $(10,10,10)$
& $(15,15,10)$
& $(20,20,20)$ \\
\midrule
DM  & 0.39   & 4.2   & 13.04  & 601.12 & $>600$ \\
GH  & $<0.01$ & $<0.01$ & 0.3  & 0.5  & 0.9\\
AGH & 0.0149   & 0.113   & 0.57   & 1.09   & 2.3 \\
\bottomrule
\end{tabular}
\vspace{-0.5cm}
\end{table}

\section{Conclusion}
\label{sec:conclusion}
This paper investigated joint model selection, GPU provisioning, parallelism configuration, and workload routing for cloud-side LLM inference under coupled latency, accuracy, memory, compute, storage, and budget constraints. We formulated the problem as a mixed-integer linear program and developed two constraint-aware heuristics, GH and AGH, that explicitly preserve feasibility throughout the allocation process. Experimental results on Azure-trace-calibrated workloads show that AGH closely matches the exact MILP solution on instances where the solver terminates, while scaling to substantially larger instances with runtime below three seconds and a speedup of at least $260\times$ relative to the MILP time limit. Under demand uncertainty and real trace replay, the proposed heuristics maintain lower cost and SLO violations than both the exact MILP and representative heuristic baselines, demonstrating the value of preserving feasibility margins in large-scale LLM-serving systems. The low runtime further enables rolling re-optimization at operational timescales, making the framework suitable for dynamic cloud resource management.

\noindent\textbf{Future work} includes a load-dependent queueing term that extends the planning-layer delay model toward engine-level dynamics, integration with a concrete serving engine (e.g., vLLM/DistServe) for closed-loop deployment, and carbon-intensity-aware tier costs.

\bibliographystyle{IEEEtran}

\end{document}